\documentclass{article}
\usepackage{acra}
\usepackage{graphicx}
\usepackage{amsmath}
\usepackage{booktabs} % For formal tables
\usepackage{multirow} % For multirow cells
\usepackage{array} % For custom column width
\usepackage{subcaption}
\usepackage{url}
\usepackage{amssymb}
\usepackage{diagbox}
\usepackage{cite}

\usepackage[table]{xcolor}
\usepackage[colorinlistoftodos]{todonotes}
\usepackage[colorlinks=true, allcolors=blue]{hyperref}
\usepackage{acronym}
\usepackage{widetable}
\usepackage{titlesec}
\usepackage{caption}

\definecolor{lightgreen}{RGB}{220,255,220} % 中等饱和度的浅绿色
\definecolor{lightblue}{RGB}{220,240,255}  % 中等饱和度的浅蓝色
\definecolor{lightyellow}{RGB}{255,255,220} % 中等饱和度的浅黄色
\definecolor{lightgray}{RGB}{240,240,240}   % 中等饱和度的浅灰色

\title{SydneyScapes: Image Segmentation for Australian Environments}
\author{Hongyu Lyu$^1$, Julie Stephany Berrio$^1$, Mao Shan, Stewart Worrall \\ The University of Sydney, Australian Centre for Robotics \\  \{h.lyu, j.berrio, m.shan, s.worrall\}@acfr.usyd.edu.au
\thanks{This research is funded by Transport for New South Wales (TfNSW), iMOVE CRC and supported by the Cooperative Research Centres program, an Australian Government initiative.}
\thanks{$^1$ Equal contribution}}

\begin{document}

\maketitle
\begin{abstract}

Autonomous Vehicles (AVs) are being partially deployed and tested across various global locations, including China, the USA, Germany, France, Japan, Korea, and the UK, but with limited demonstrations in Australia. The integration of machine learning (ML) into AV perception systems highlights the need for locally labelled datasets to develop and test algorithms in specific environments. To address this, we introduce SydneyScapes—a dataset tailored for computer vision tasks of image semantic, instance, and panoptic segmentation. This dataset, collected from Sydney and surrounding cities in New South Wales (NSW), Australia, consists of 756 images with high-quality pixel-level annotations. It is designed to assist AV industry and researchers by providing annotated data and tools for algorithm development, testing, and deployment in the Australian context. Additionally, we offer benchmarking results using state-of-the-art algorithms to establish reference points for future research and development. The dataset is publicly available at \href{https://hdl.handle.net/2123/33051}{https://hdl.handle.net/2123/33051}.

\end{abstract}

\section{Introduction}

Perception systems for AVs have evolved from pure algorithmic methodologies to (ML) methods. While ML models usually outperform classical methods, they require large amounts of data to be trained \cite{10415295}. For example, the models trained in data from Germany can generalise in an environment within the same domain. Nevertheless, when the same model is deployed in a different location, the capability for networks to generalise is reduced as this is new data that has never been seen before in the training stage \cite{Zhang_2017_ICCV}. This is the case of ML models that use image data as input to infer the corresponding semantic, instance or panoptic labels. 

\begin{figure}[t!]
\centering
\begin{subfigure}[]{0.48\columnwidth}
\centering
\includegraphics[trim={20cm 11cm 10cm 12cm},clip, width=\columnwidth]{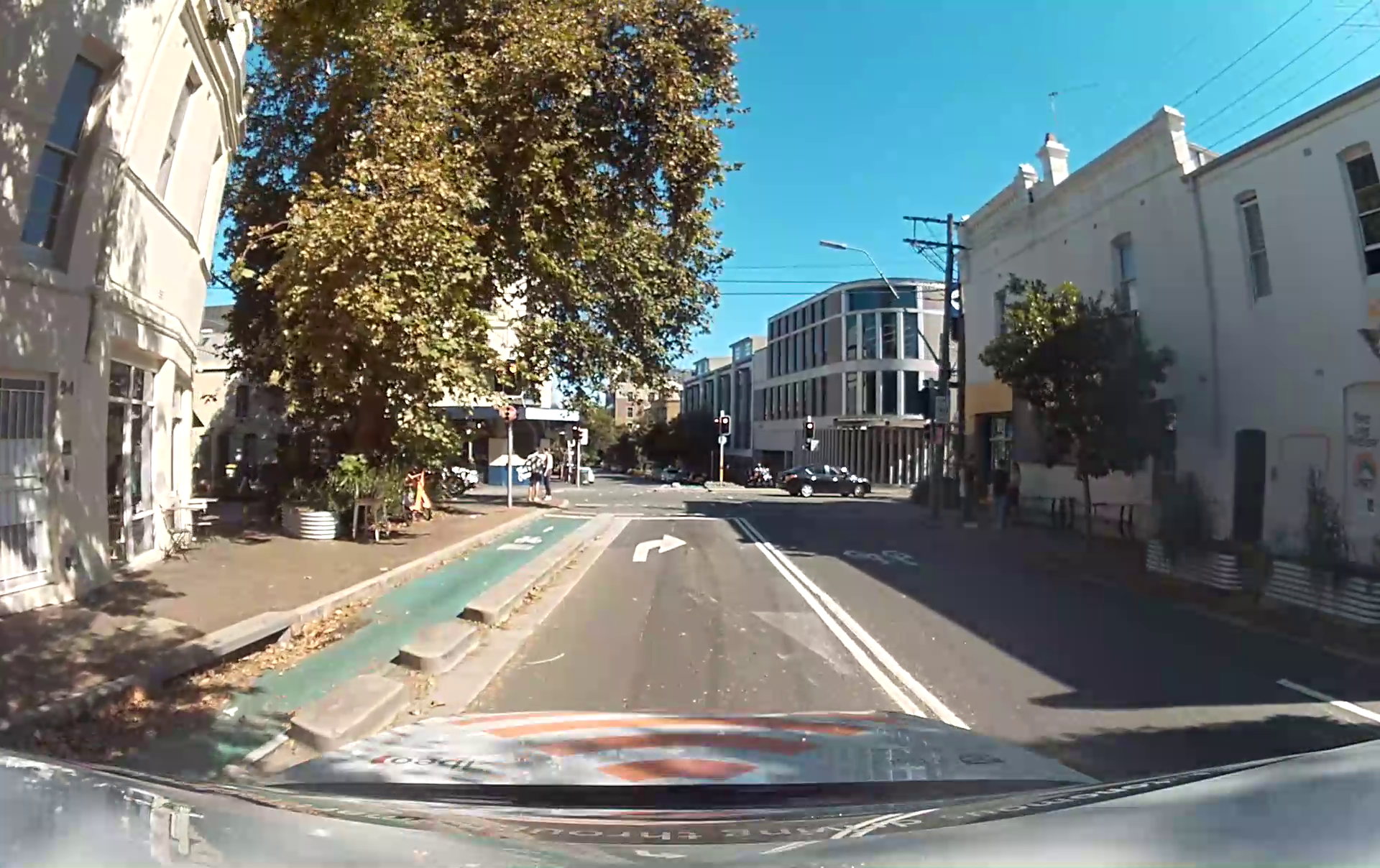}
\caption{\small Original image}
\label{fig:original}
\end{subfigure}
\begin{subfigure}[]{0.48\columnwidth}
\centering
\includegraphics[trim={20cm 11cm 10cm 12cm},clip, width=\columnwidth]{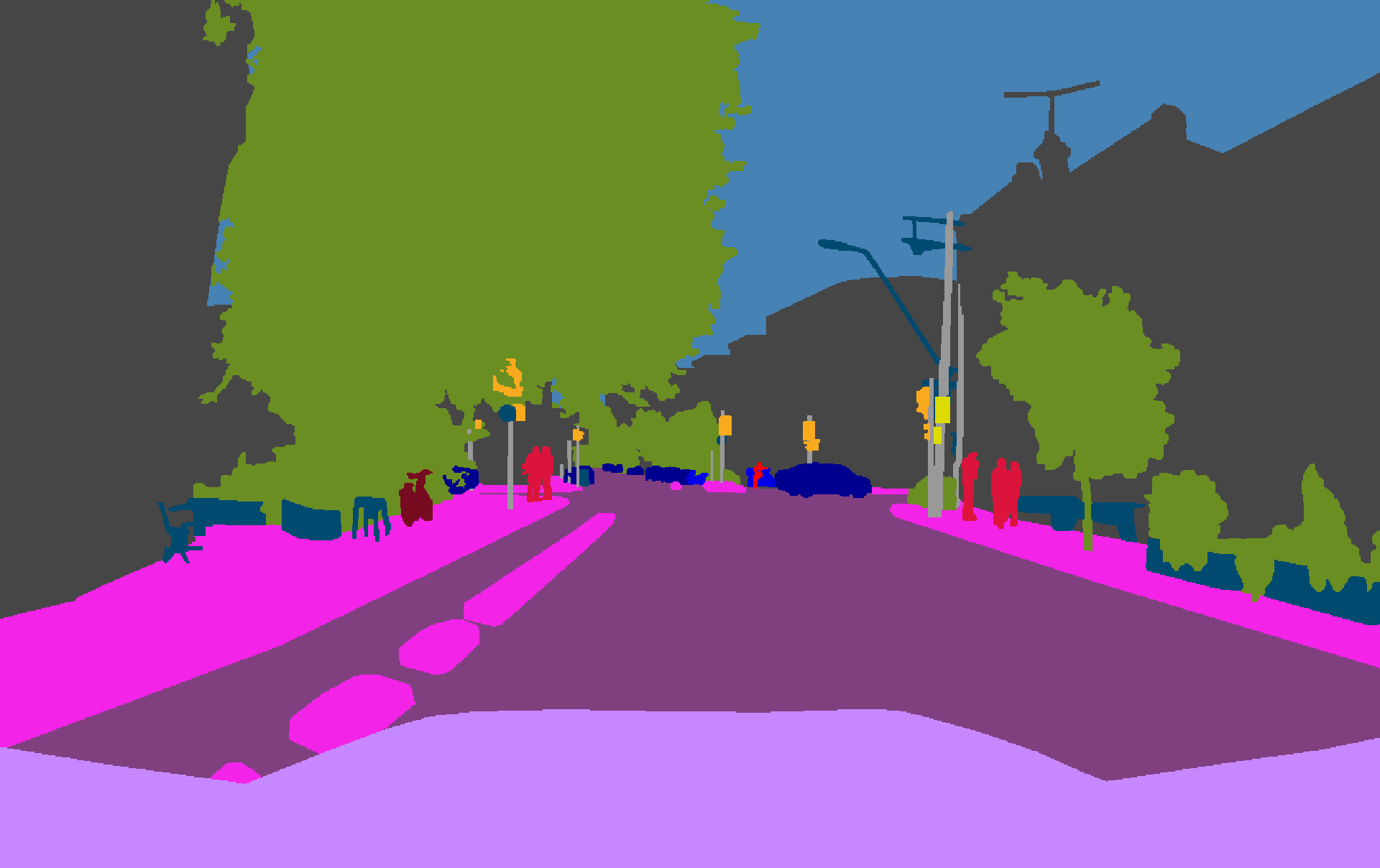}
\caption{\small Semantic segmentation}
\label{fig:semantic}
\end{subfigure}
\begin{subfigure}[]{0.48\columnwidth}
\centering
\includegraphics[trim={20cm 11cm 10cm 12cm},clip, width=\columnwidth]{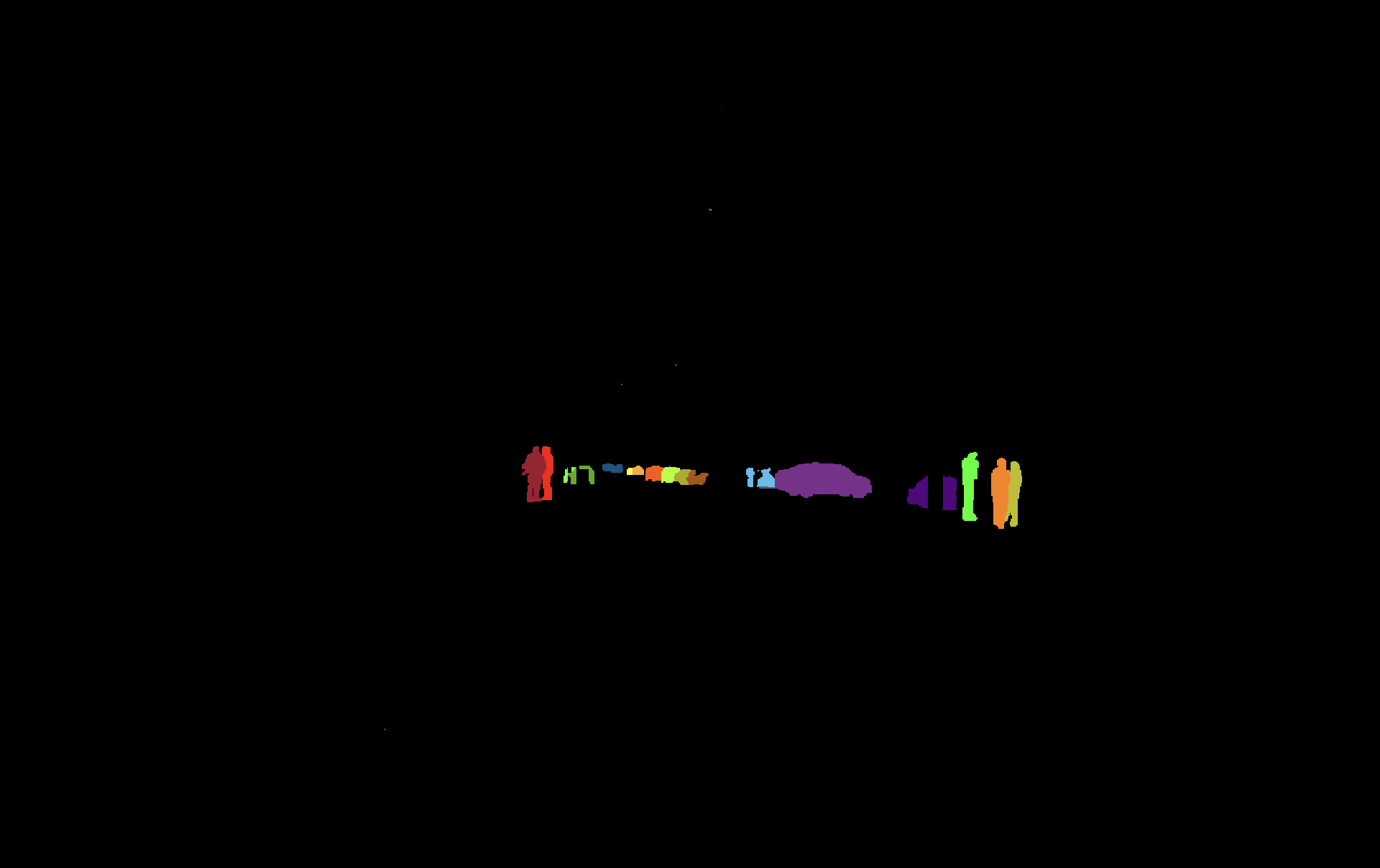}
\caption{\small Instance segmentation}
\label{fig:instance}
\end{subfigure}
\begin{subfigure}[]{0.48\columnwidth}
\centering
\includegraphics[trim={20cm 11cm 10cm 12cm},clip, width=\columnwidth]{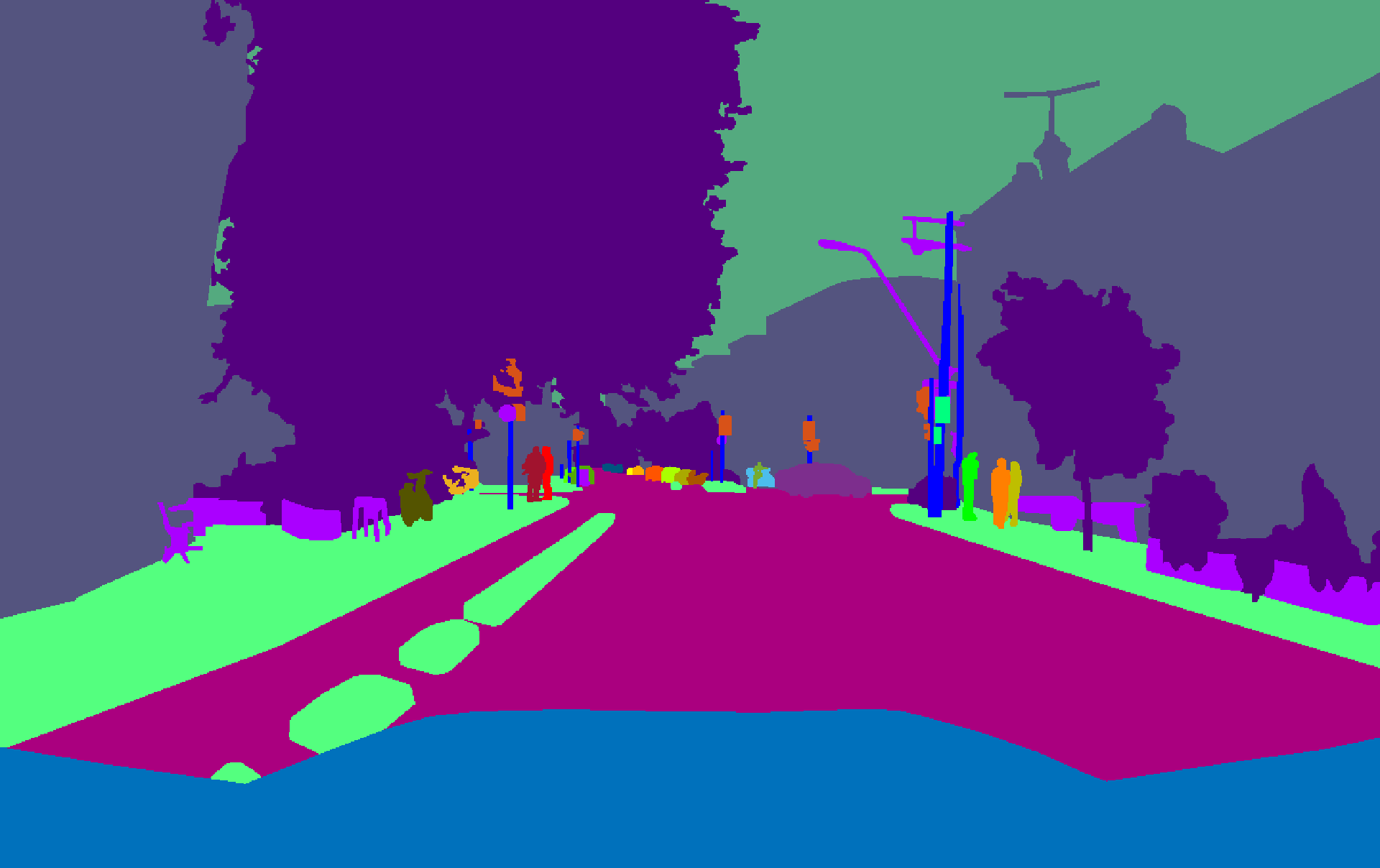}
\caption{\small Panoptic segmentation}
\label{fig:panoptic}
\end{subfigure}
\caption{\small   Segmentation techniques applied to an urban scene, Fig. \ref{fig:semantic} shows a city street, Fig. \ref{fig:semantic} categorises areas by class using different colours. Fig. \ref{fig:instance} highlights individual objects with unique colours and Fig. \ref{fig:panoptic} identifies objects by both class and instance.}
\label{fig:semantic-instance}
\end{figure} 

Semantic segmentation is the computer vision task of dividing an image into semantically meaningful regions. For example, in a scene containing a person and a car, semantic segmentation would enable the perception system to distinguish between the person and the car \cite{feng2020deep}. Instance segmentation goes one step further by also identifying individual instances of objects within each region. For example, in the scene containing people and cars, instance segmentation allows to identify the number of people and vehicles in the picture. Finally, panoptic segmentation merges the tasks of semantic and instance segmentation, into one. The panoptic segmentation task classifies the pixels in the image as belonging to a class label yet also identifies what instance of that class they belong to. Fig. \ref{fig:semantic-instance} shows an original image with corresponding semantic, instance, and panoptic segmentation. 

The topic of domain adaptation focuses on mitigating performance degradation when ML models are applied to new testing domains \cite{Chen_Dou_Chen_Qin_Heng_2019}. In the context of ML for image segmentation, domain shifts can occur due to variations in illumination, weather conditions, or changes in the sensor or environment, such as testing on images from a different country. 
To address this issue, numerous domain adaptation approaches focuses on adapting models trained on a source domain to target domains with unlabeled data \cite{Zhao_2024_WACV}. The source domain refers to the dataset where the ML model is initially trained and the target domain is the new dataset environment where the model will be applied after training. 
To verify or test the effectiveness of the domain adaptation methods, labeled data from the target domain is necessary.

This paper introduces SydneyScapes, a dataset designed for semantic, instance, and panoptic segmentation of urban environments. Collected in New South Wales, Australia, under different illumination and crowd conditions, using a front-facing camera on an urban vehicle, SydneyScapes offers a resource customised to the characteristics of the city of Sydney and surroundings. The dataset aims to contribute to the research community with the data and tools to test, refine, and deploy their technologies in Australian environments, and promote research in urban scene understanding and adaptation to local conditions.

The contributions of this paper are as follows:
\begin{itemize}
    \item We present an image segmentation dataset, with ground truth labels for semantic, instance, and panoptic segmentation.
    \item We introduce a cloud-based visualisation tool designed to interactively explore the dataset.
    \item We establish a benchmark for evaluating segmentation performance within Australian contexts. 
\end{itemize}

\section{Background}

The development of datasets for tasks like semantic, instance, and panoptic segmentation has allowed the advance of computer vision in urban environments. While all these datasets have contributed to global research, there has been a notable gap in data representing the unique characteristics of Australian urban landscapes.

Research in computer vision for AVs has seen significant growth since the release of the KITTI dataset \cite{Geiger2013IJRR}, which provided researchers with the data and tools to develop and test algorithms for perception and other aspects of AV operations.
The Cityscapes dataset \cite{Cordts2016Cityscapes} was a pioneer for urban scene understanding, it contains annotated images from various cities across Germany.
Building on the foundation of KITTI dataset, the work in \cite{Weber2021NEURIPSDATA} and \cite{vip_deeplab} extended its utility with semantic annotations and video panoptic segmentation for the original images. This enriched the original dataset, enabling research in semantic segmentation and other tasks in AV perception.
Later, KITTI-360 \cite{Liao2022PAMI} and Audi A2D2 \cite{geyer2020a2d2} were introduced as two new German-based dataset that provides annotations for both semantic and instance segmentation, extending beyond 2D images to include 3D point clouds as well.

Datasets featuring semantic, instance, and panoptic segmentation have gained popularity globally due to their practicality for local deployment. Examples include the Berkeley Deep Drive dataset \cite{bdd100k}, collected across various U.S. cities, ApolloScape \cite{wang2019apolloscape}, which offers car instance and lane segmentation data from China, and the nuScenes dataset \cite{nuscenes2019}, collected in Singapore. Additionally, the Indian Driving Dataset \cite{8659045} focuses on the unique challenges of Indian roads, while the Carl-d dataset \cite{BUTT2022116667} from Pakistan captures the complexities of urban traffic.

Mapillary Vistas \cite{8237796} attempts to address the geographical limitations of datasets by capturing images from diverse locations worldwide. However, the dataset's global scope, combined with data sourced from various devices and viewpoints, results in a lack of concentrated focus on specific Australian regions and the perspectives unique to individual vehicles. Other approaches have also been explored, such as using synthetic images to introduce a wider variety of environments, including examples like \cite{Ros_2016_CVPR}. Nevertheless, the domain problem persists when it comes to applicability/deployment. 

When focusing on Australian-specific landscapes, more recently, \cite{vidanapathirana2023wildscenes} presented a semantically labelled dataset of natural and unstructured environments collected in two distinct locations in Australia. 
Australia’s cities have distinct features like native flora, road markings, and unique architectural styles that differs from those in Europe, Asia, or North America. The SydneyScapes dataset addresses this gap with a comprehensive collection of images from urban environments in New South Wales, Australia. SydneyScapes includes annotations for semantic, instance, and panoptic segmentation, covering various environmental conditions, such as different times of day and weather scenarios. This enables research in domain adaptation and model robustness in Australian urban environments. With this dataset, we aim to represent the nuances of Australian environments, making it a resource for training and testing segmentation models intended for use in Australia.

\section{The SydneyScapes Dataset}

The data collection, post processing and annotation of images for semantic, instance, and panoptic segmentation is an important step in training and testing a computer vision system to accurately recognise and classify objects in images from our local environment. This section explains the process from the sensor setup to the final labelled dataset. 

 \subsection{Data Collection}

\begin{figure}[t!]
\centering
\begin{subfigure}[]{\columnwidth}
\centering
\includegraphics[width=\columnwidth]{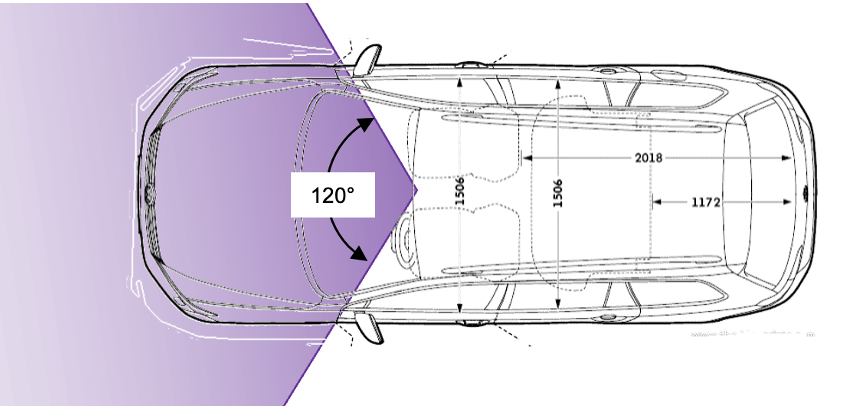}
\caption{\small Data collection platform, top-down view.}
\label{fig:topdown}
\end{subfigure}
\begin{subfigure}[]{\columnwidth}
\centering
\includegraphics[width=\columnwidth]{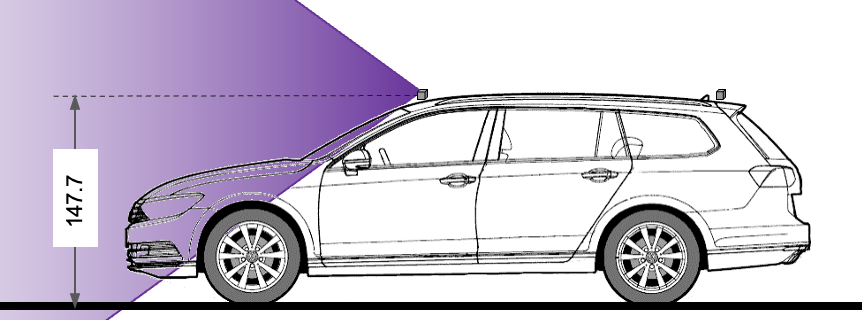}
\caption{\small Data collection platform, side view. Units are in cm.}
\label{fig:side}
\end{subfigure}

\caption{\small Camera's positioning and coverage area for data collection.}
\label{fig:platform}
\end{figure}

For data collection, we used a single camera mounted on an urban vehicle (Volkswagen Passat). The camera, an SF3324 automotive GMSL model, is equipped with an ONSEMI CMOS Image Sensor AR0231 (2M Pixel) and a SEKONIX ultra-high-resolution lens. The lenses provide a $120^{\circ}$ horizontal field of view (FOV) and a $73^{\circ}$ vertical FOV. Images were captured at a resolution of 1928 x 1208 pixels (2.3M pixels). As shown in Fig. \ref{fig:platform}, the camera was mounted horizontally and centered on the data collection platform at a height of 1.47 meters above the ground.

The data collection was primarily conducted in the city of Sydney, Australia, with some of the images also captured in nearby towns. The data was recorded in a naturalistic manner, with the vehicle being driven on the road as it would be under normal conditions. The focus was on capturing a variety of scenarios across different lighting conditions and diverse crowd densities. From multiple driving sessions, we selected a total of 756 images for labeling. These images were chosen to represent different scenarios for training and evaluating segmentation models in Australian environments.

For visual consistency, we divided the dataset into three parts—SydneyScapes Day, SydneyScapes Night, and SydneyScapes People—with the objective of fine-tuning ML methods on a diverse range of data, including both day and night conditions, as well as different types of objects and scenes. 

\begin{figure}[t!]
\centering

% 第一组图片 - Daytime
\begin{subfigure}[]{0.98\columnwidth}
\centering
\includegraphics[width=0.48\columnwidth]{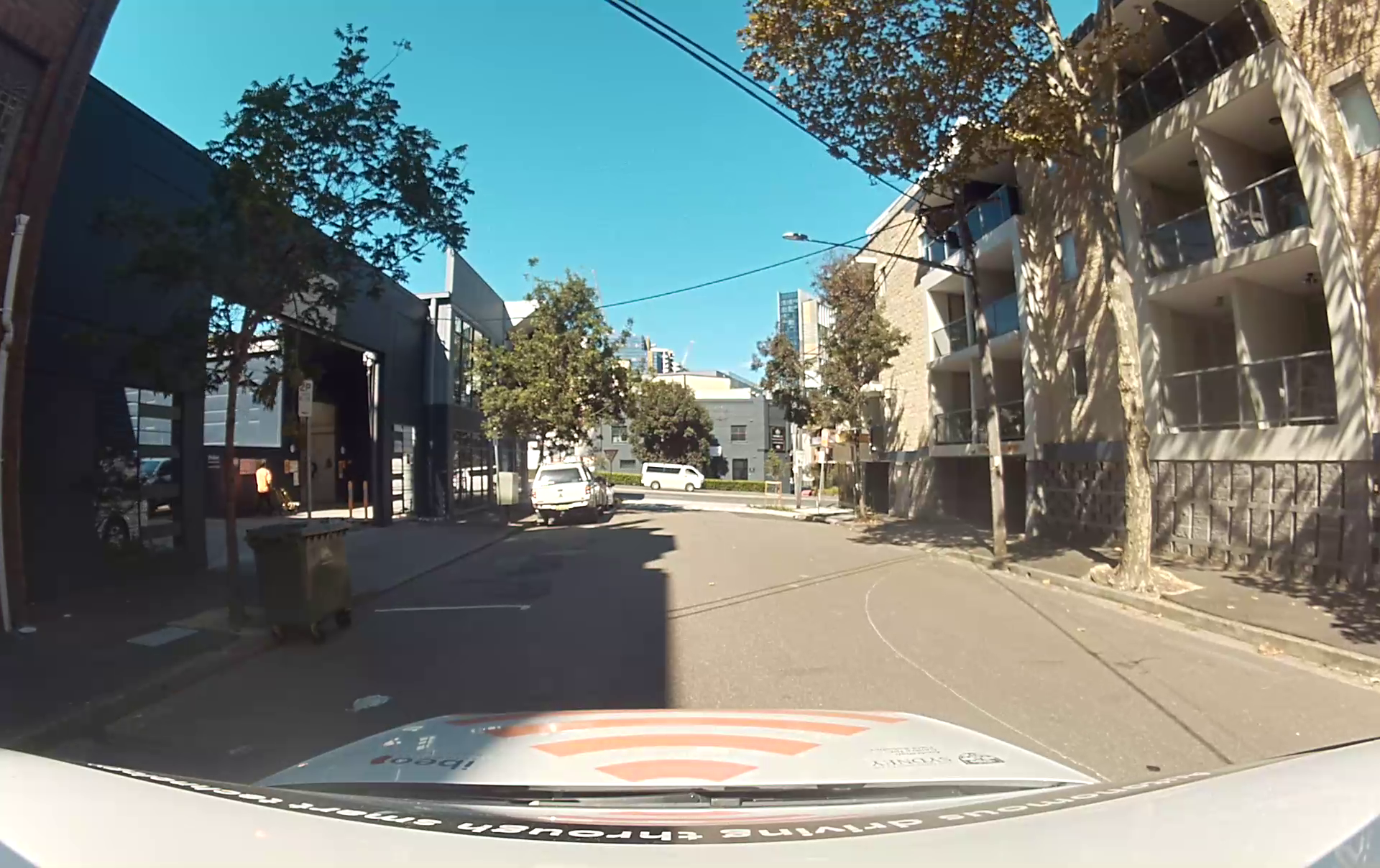}
\includegraphics[width=0.48\columnwidth]{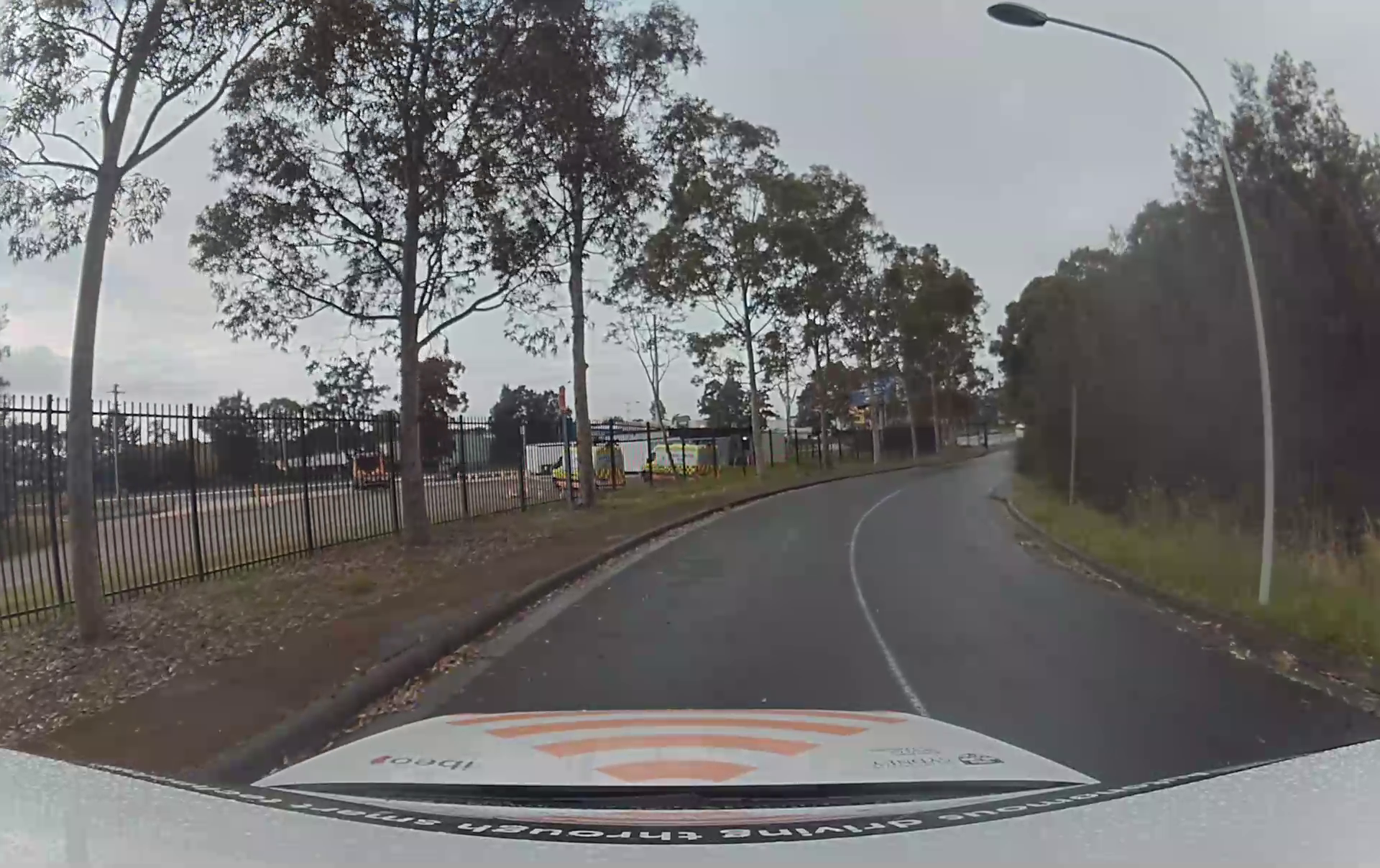}
\caption{\small Daytime image captures in different locations.}
\label{fig:NSW_dataset}
\end{subfigure}

% 第二组图片 - Nighttime
\begin{subfigure}[]{0.98\columnwidth}
\centering
\includegraphics[width=0.48\columnwidth]{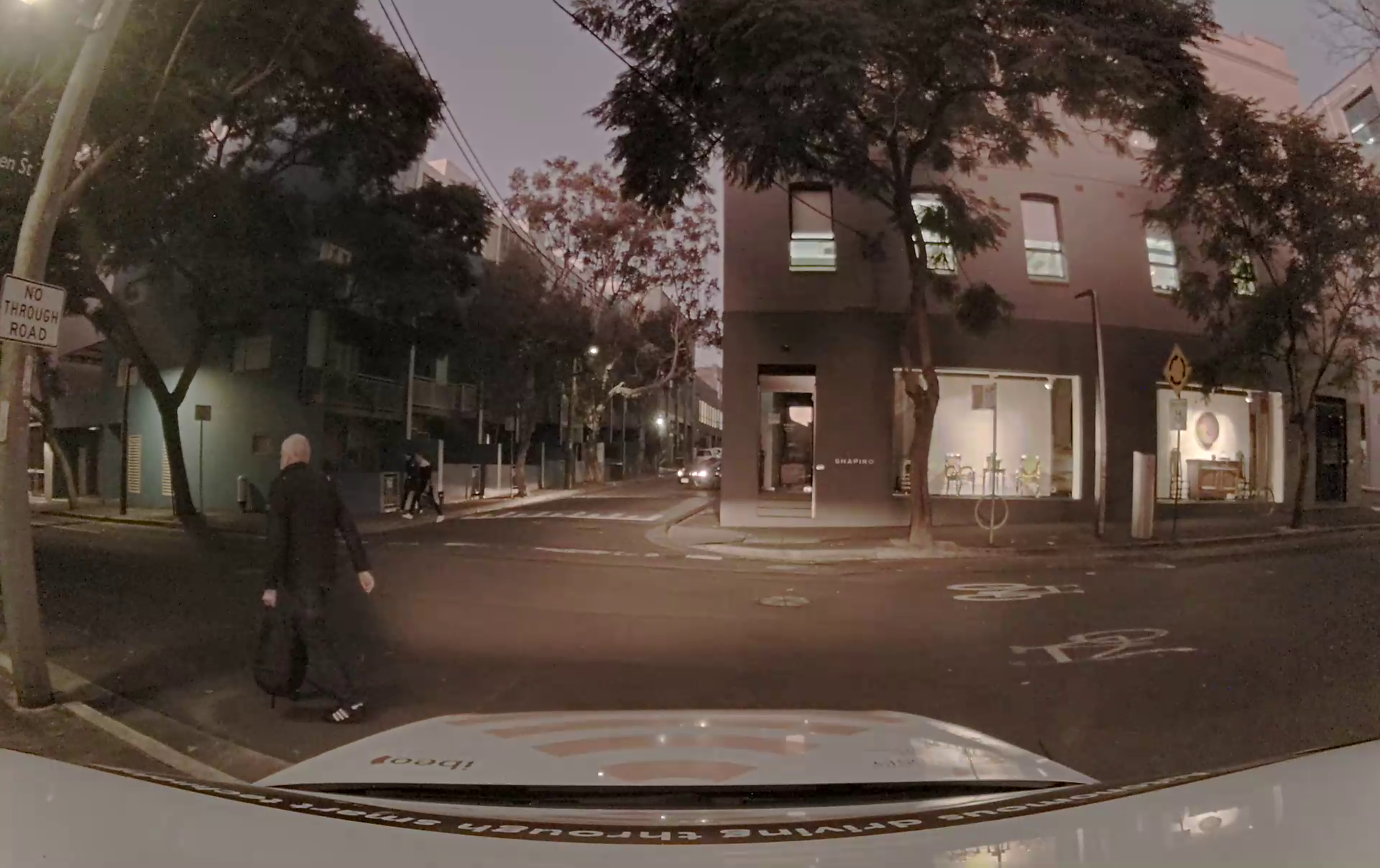}
\includegraphics[width=0.48\columnwidth]{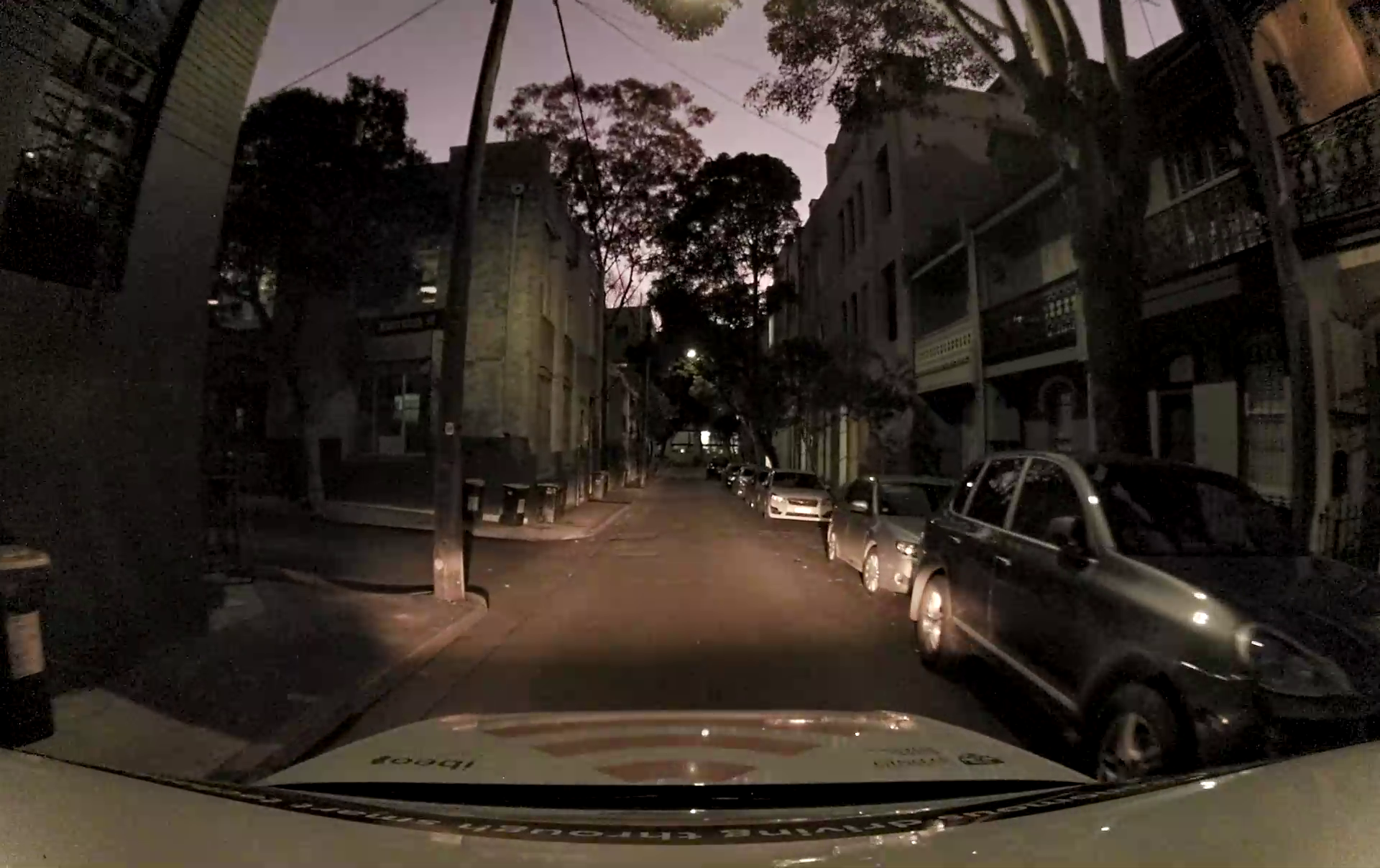}
\caption{\small Nighttime images captures in different locations.}
\label{fig:NSW_dataset_night}
\end{subfigure}

% 第三组图片 - Crowded locations
\begin{subfigure}[]{0.98\columnwidth}
\centering
\includegraphics[width=0.48\columnwidth]{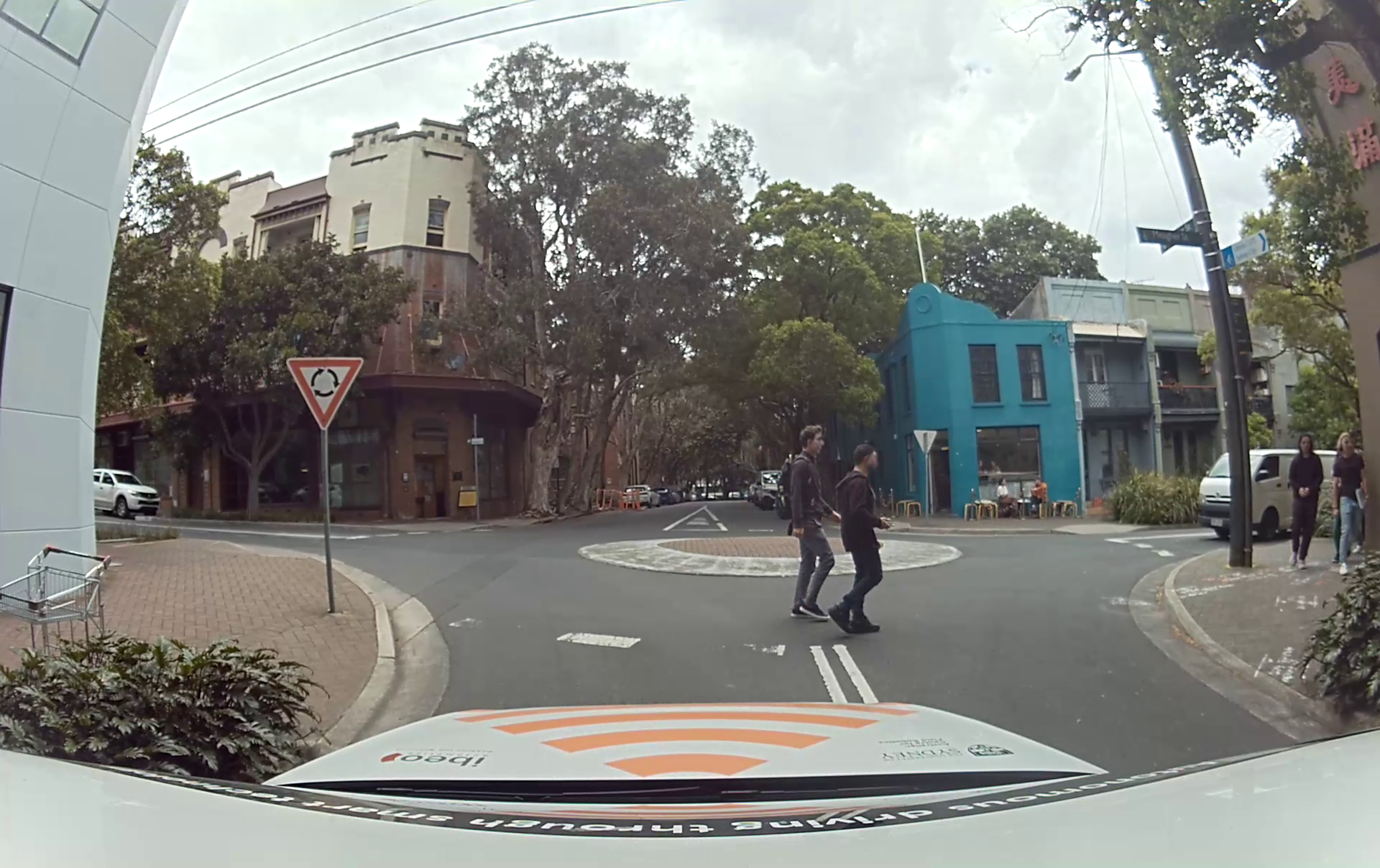}
\includegraphics[width=0.48\columnwidth]{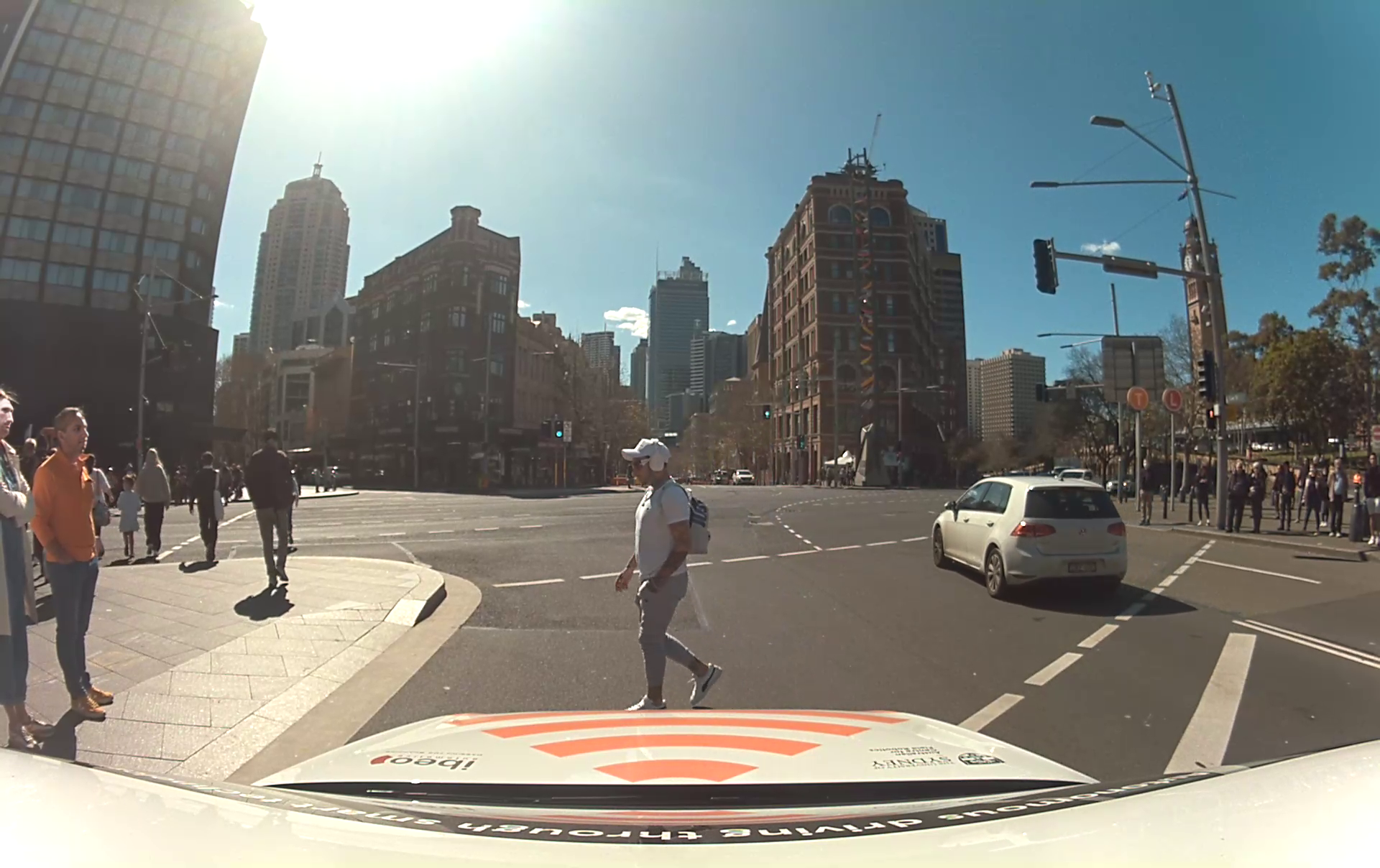}
\caption{\small Images captures in different crowded locations.}
\label{fig:NSW_dataset_people}
\end{subfigure}

\caption{\small Sample images from the SydneyScapes dataset: (a) Day subset, (b) Night subset, and (c) People subset.}
\label{fig:NSW_dataset_combined}

\end{figure}

\subsubsection{SydneyScapes Day}
This part of the dataset includes 332 images of different cities around NSW, including Cudal, Orange and Sydney. The data collection took place on rural roads, highways and urban areas. This dataset subset comprises several conditions such as: sunny, cloudy, rainy, strong shadows, etc., as shown in Fig. \ref{fig:NSW_dataset}.

\subsubsection{SydneyScapes Night}

Changes in illumination can challenge CV tasks by altering the appearance of objects, making recognition difficult.  We recorded data at night time to assess image segmentation algorithms under low-light conditions. This dataset subset comprises 104 images collected in urban areas of Sydney, as shown in Fig. \ref{fig:NSW_dataset_night}.

\subsubsection{SydneyScapes People}

This third part of the dataset focuses on detecting and segmenting people in urban driving environments. Accurate pedestrian detection is important for safe navigation, especially in high-density urban areas and zones with increased pedestrian presence. This dataset subset includes 320 images collected in different suburbs of Sydney, as shown in Fig. \ref{fig:NSW_dataset_people}.

\subsection{Anonymisation} 

In compliance with local authority policies requiring data to be anonymised before publication, we developed a post-processing pipeline with two key algorithms: one to remove and replace human faces and the other to blur number plate information.

\begin{figure}[t!]
\centering
\begin{subfigure}[]{0.48\columnwidth}
\centering
	\includegraphics[trim={0cm 10cm 23cm 3cm},clip ,width=\columnwidth]{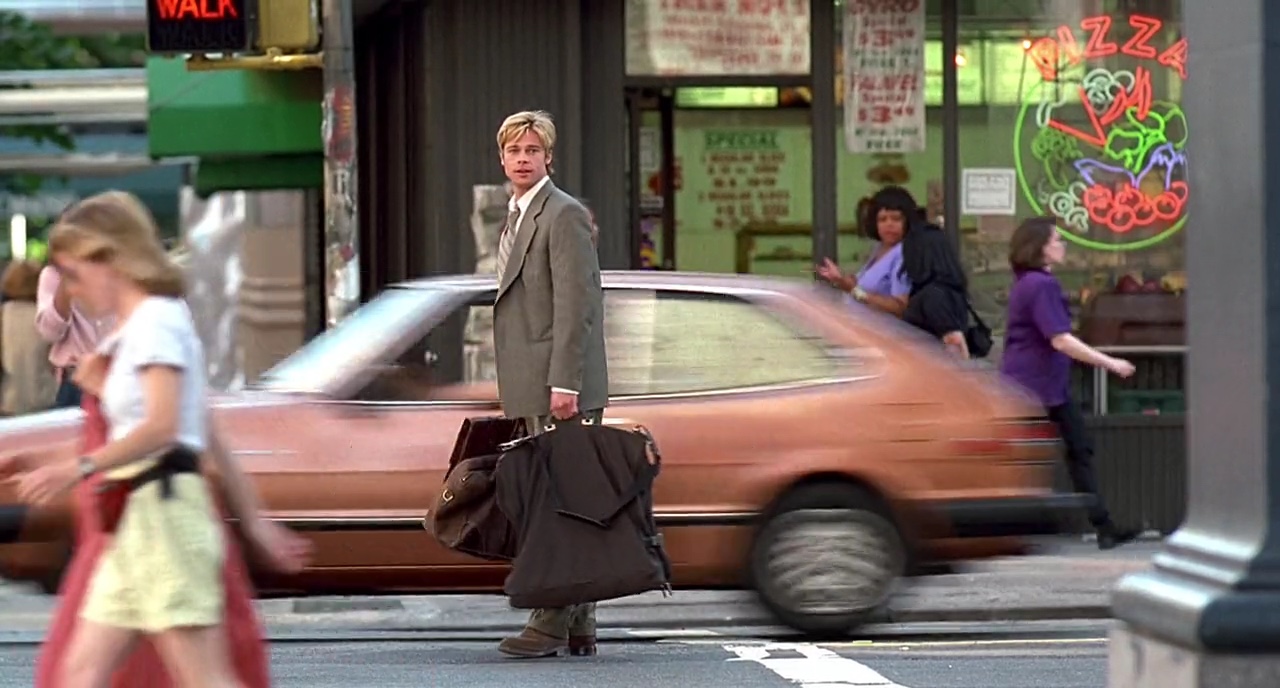}
    \caption{\small Original image}
    \end{subfigure}
\begin{subfigure}[]{0.48\columnwidth}
\centering
	\includegraphics[trim={45.1cm 10cm 23cm 3cm},clip, width=\columnwidth]{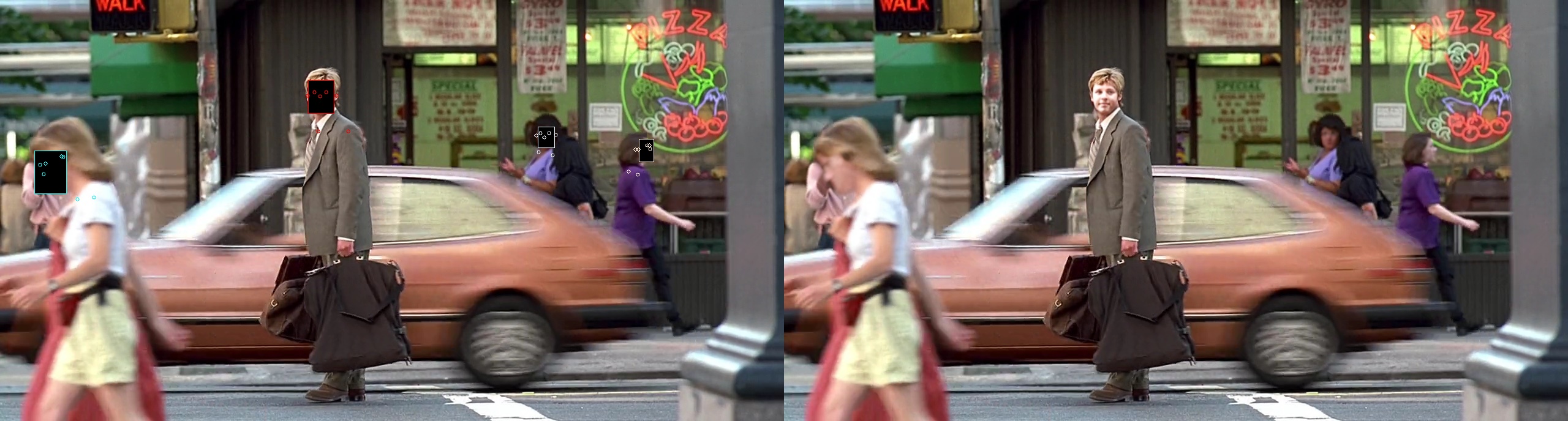}
    \caption{\small Face anonymisation}
    \end{subfigure}

\begin{subfigure}[]{0.48\columnwidth}
\centering
	\includegraphics[trim={7cm 0cm 7cm 5.5cm},clip,width=\columnwidth]{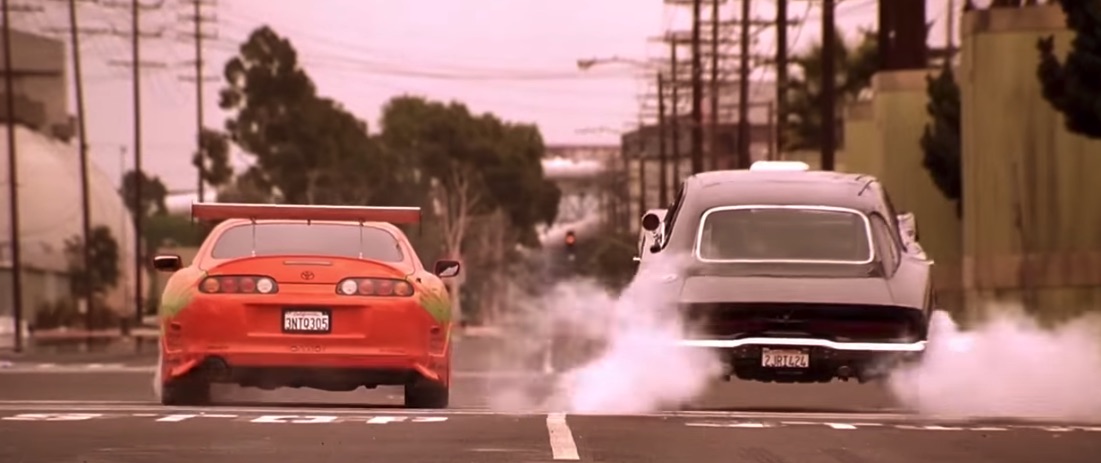}
    \caption{\small Original image}
    \end{subfigure} 
\begin{subfigure}[]{0.48\columnwidth}
\centering
	\includegraphics[trim={5cm 0cm 5cm 4cm},clip,width=\columnwidth]{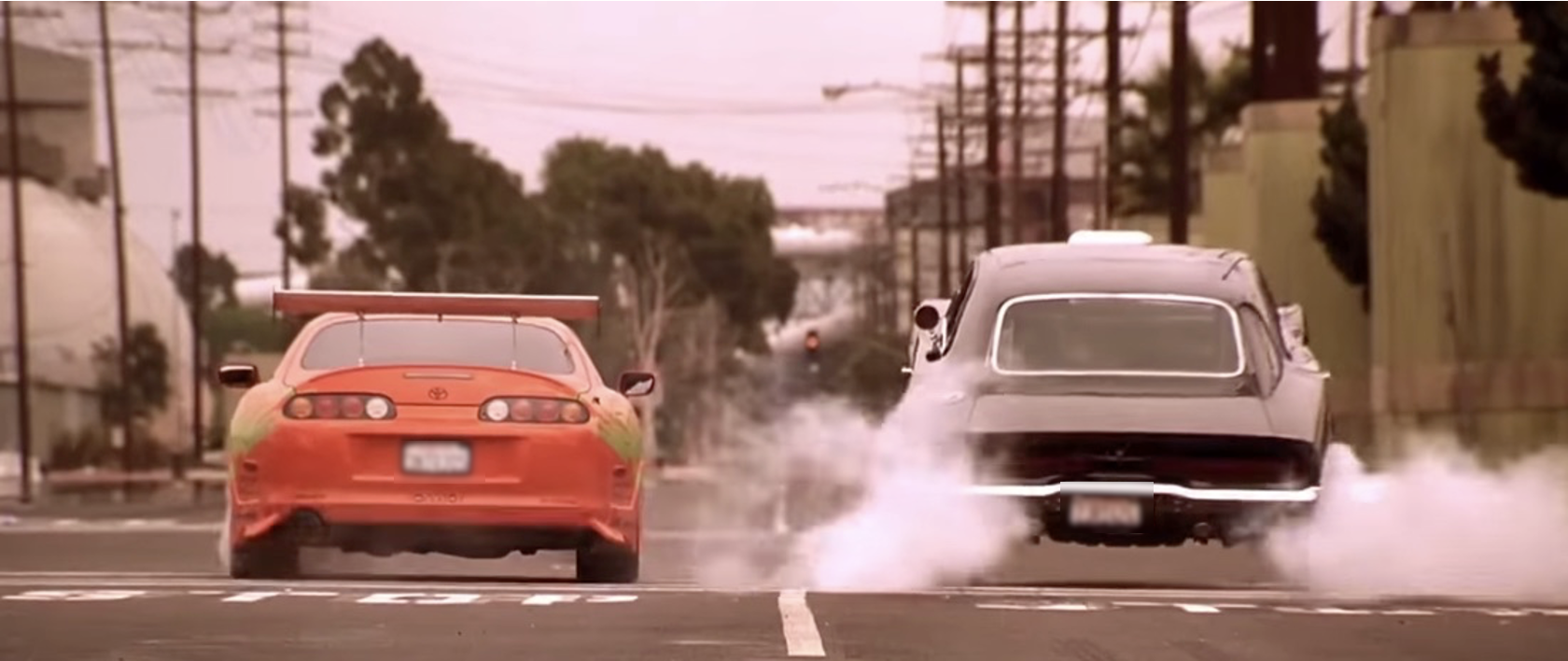}
    \caption{\small No. plate anonymisation}
    \end{subfigure}
\caption{\small  Original images, which are publicly available on the internet, alongside the processed images after applying the face and number plate anonymisation.}
\label{fig:anonymisation}
\end{figure}

For face anonymisation, we used the DeepPrivacy algorithm \cite{Hukkels2019DeepPrivacyAG}, which anonymises faces by generating realistic, privacy-safe substitutes while preserving the original background of the image. To anonymise number plates, we employed the DashCamCleaner implemented in \cite{dashcamcleaner}, this method uses YOLOv8 \cite{yolov8} algorithm to detect the bounding boxes containing number plates. Then, it applied a Gaussian blur to the pixels within these bounding boxes. Figure \ref{fig:anonymisation} shows in the first column the original images with visible faces and number plates. The second column shows the anonymised version where the faces have been replaced and the number plates have been blurred to maintain privacy.

\subsection{Labelling} 

\subsubsection{Labels}
For labelling the images, we drew inspiration from the Cityscapes dataset \cite{Cordts2016Cityscapes}, adopting their labeling policy and original label classes. We reorganised the class definitions into groups similar. The dataset labels are divided into seven groups: Flat, Human, Vehicle, Construction, Object, Nature, and Void. Each group contains specific labels for various items. For instance, the ``Vehicle" group includes labels for several types of transportation, such as ``Car", ``Bus", ``Truck", ``On rails", ``Motorcycle", ``Bicycle", ``Caravan", and ``Trailer". Only the ``Human" and ``Vehicle" groups are classified as ``things", for which we have instance-level annotations. A description of the groups, instance classes, and specific labels is shown in the table \ref{tab:labels}.

To train the ML models, we allocate 80\% of the data from each dataset for training and reserve the remaining 20\% for validation. This approach ensures that the ML model is trained on a representative subset of the data, while a separate portion is used to evaluate model performance. Consequently, this method results in a split of 594 training images and 162 validation images.

\begin{table}[t!]
\centering
\caption{\small Labels used in the dataset for image segmentation annotations.}
\resizebox{\columnwidth}{!}{%
\begin{tabular}{|l|l|l|}
\hline
\rowcolor[HTML]{C0C0C0} 
\textbf{Group} & \begin{tabular}[c]{@{}l@{}}\textbf{Instance}\\\textbf{Class}\end{tabular}& \textbf{Classes}                                          \\ \hline
Flat & Stuff          
& \begin{tabular}[c]{@{}l@{}}Road, Sidewalk, Terrain, \\ Parking, Ground\end{tabular}        \\ \hline
Construction & Stuff  & \begin{tabular}[c]{@{}l@{}}Building, Wall, Fence,\\ Guard rail, Bridge, Tunnel \end{tabular}                  \\ \hline
Object & Stuff        & \begin{tabular}[c]{@{}l@{}}Pole, Traffic sign, Traffic light  \end{tabular}          \\ \hline
Nature & Stuff        & Vegetation, Sky                       \\ \hline
Human & Thing        & Person, Rider                                             \\ \hline
Vehicle & Thing       & \begin{tabular}[c]{@{}l@{}}Car, Truck, Bus, On rails, \\   Motorcycle, Bicycle, Caravan, \\  Trailer \end{tabular}    \\ \hline
Void & Stuff           & Dynamic, Static                                           \\ \hline
\end{tabular}
}
\label{tab:labels}
\end{table}

\subsubsection{Visualisation}

\begin{figure}[b]
    \centering
    \includegraphics[width=\columnwidth]{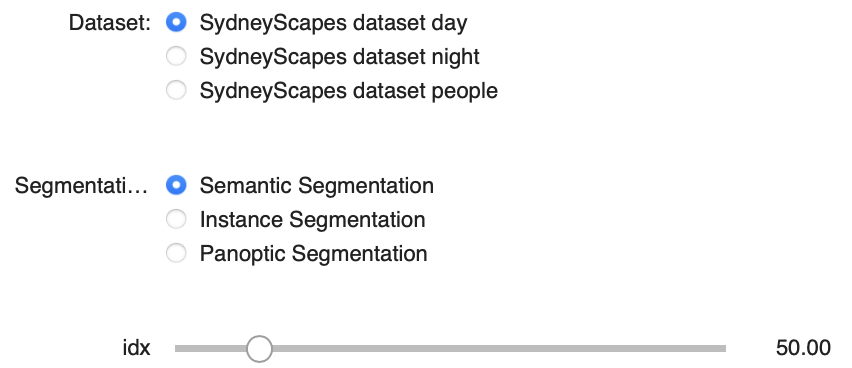}
    \caption{\small Selection panel for dataset inspection.}
    \label{fig:visual_tool}
\end{figure}

\begin{figure*}[t!]
\centering
\includegraphics[width=\textwidth]{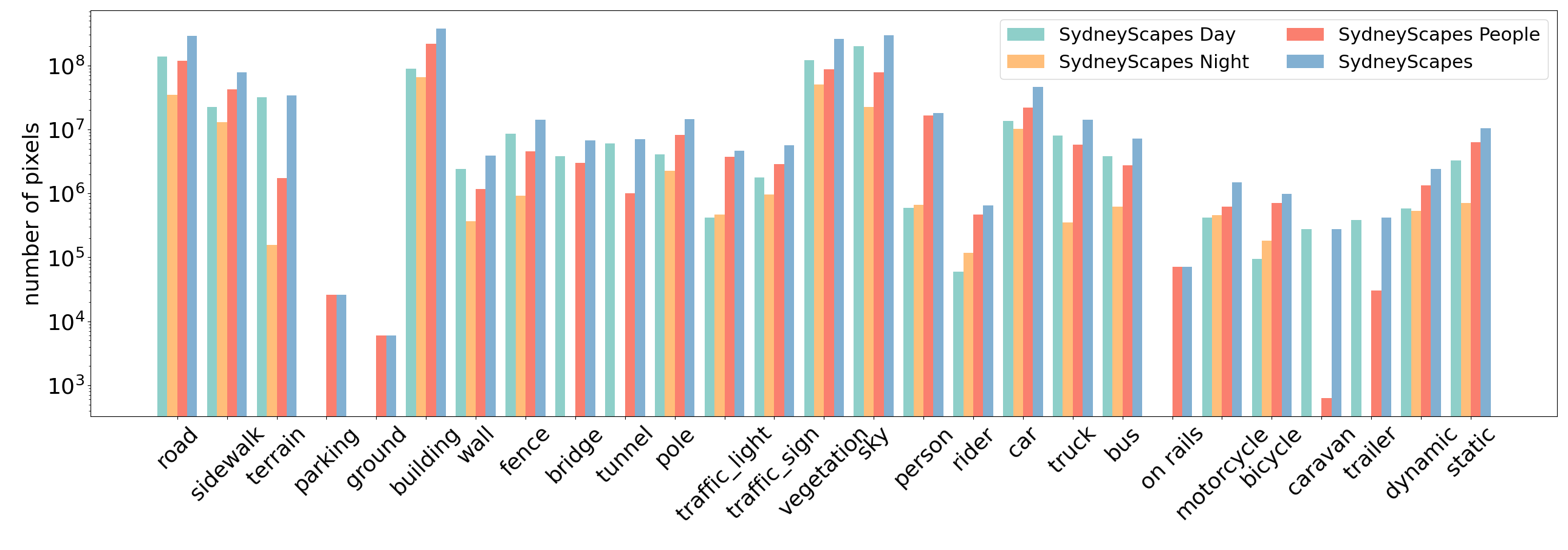}
\caption{\small Distribution of annotated pixels (y-axis) across each class (x-axis) in the SydneyScapes dataset and its subsets.}
\label{fig:semantics_distribution}
\end{figure*}

We use colab notebooks to visualise the data as it is easy for the broad community to quickly inspect the data without much experience. The dataset can be downloaded from \href{https://hdl.handle.net/2123/33051}{https://hdl.handle.net/2123/33051}, and the visualisation tool and instructions are accessible through the following link:
\url{https://colab.research.google.com/drive/1e0AYVLEzfEthHXNJi5ZKCayPKYX21D1P?usp=sharing}. Fig. \ref{fig:visual_tool} depicts the user interface for selecting and viewing different datasets and segmentation tasks. Users can choose between three datasets: ``SydneyScapes dataset day", ``SydneyScapes dataset night", and ``SydneyScapes dataset people". Additionally, there is a slider "idx" that allows users to scroll through various images within the chosen dataset.

\begin{figure}[t!]
\centering

\begin{subfigure}[]{0.98\columnwidth}
\centering
\includegraphics[trim={0cm 0 32cm 7.5cm},clip, width=0.48\columnwidth]{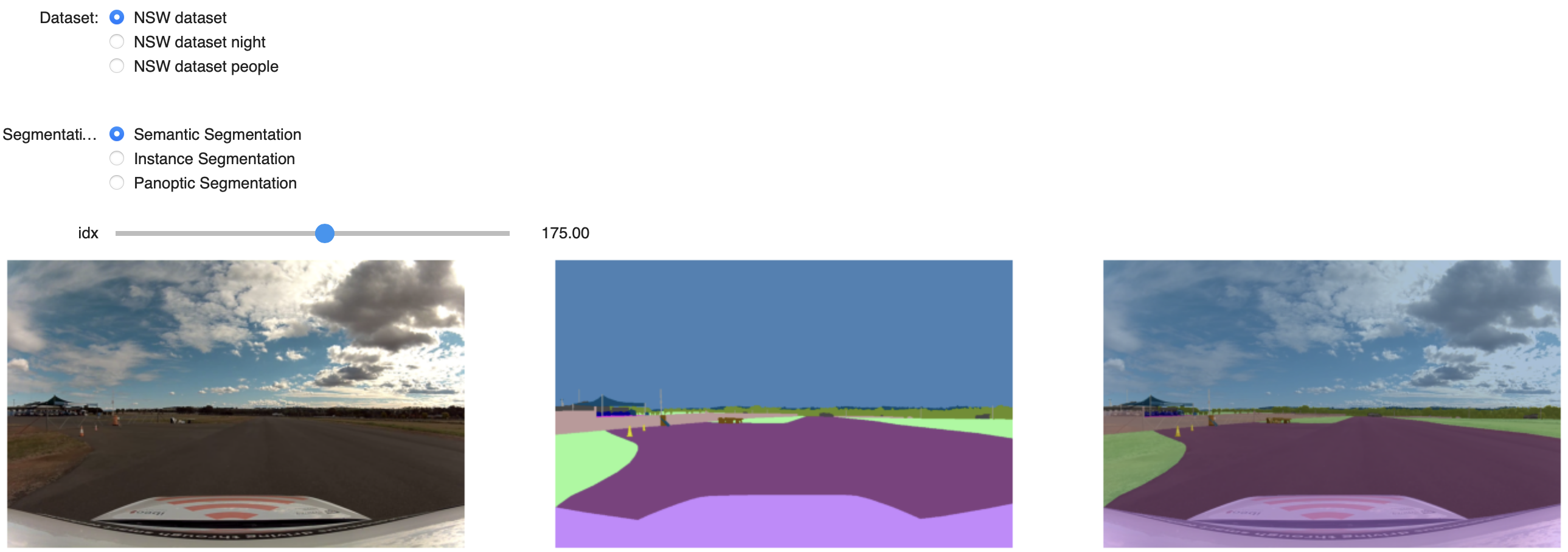}
\includegraphics[trim={32cm 0 0 7.5cm},clip, width=0.48\columnwidth]{figures/Visualisation3.png}
\caption{\small Daytime image and semantic mask}
\end{subfigure}

\begin{subfigure}[]{0.98\columnwidth}
\centering
\includegraphics[trim={0cm 0 32cm 7.5cm},clip, width=0.48\columnwidth]{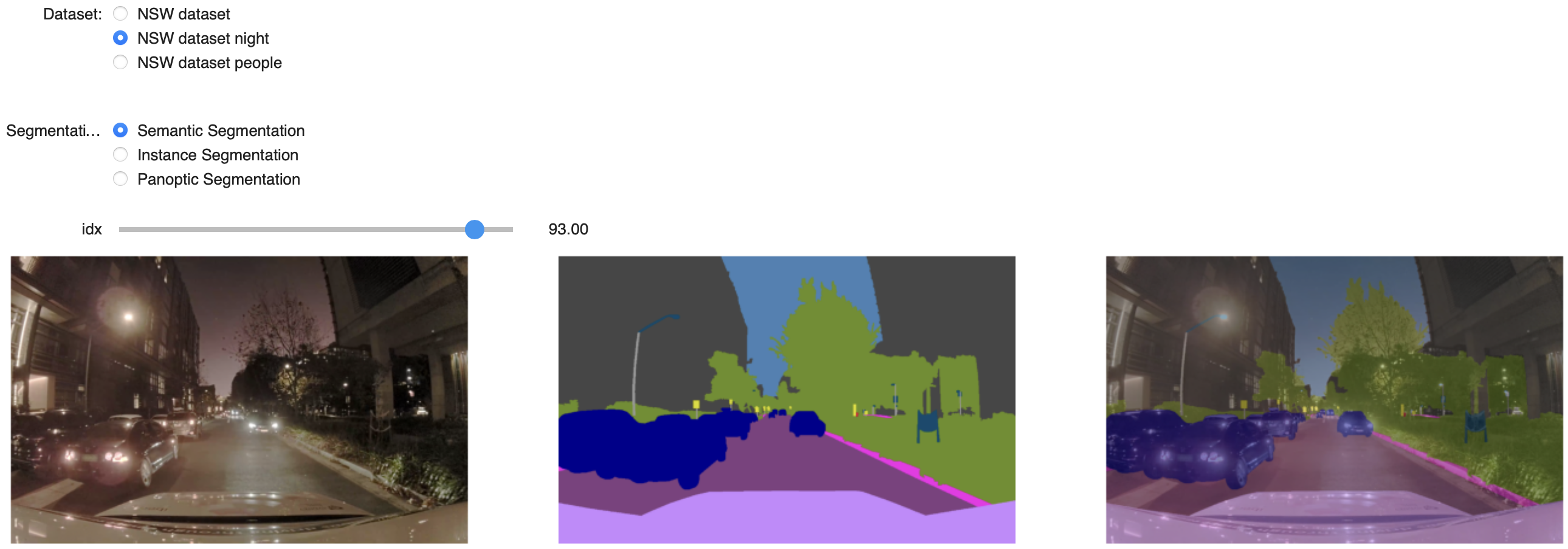}
\includegraphics[trim={32cm 0 0 7.5cm},clip, width=0.48\columnwidth]{figures/Visualisation2.png}
\caption{\small Nighttime image and semantic mask}
\end{subfigure}

\begin{subfigure}[]{0.98\columnwidth}
\centering
\includegraphics[trim={0cm 0 32cm 8cm},clip,width=0.48\columnwidth]{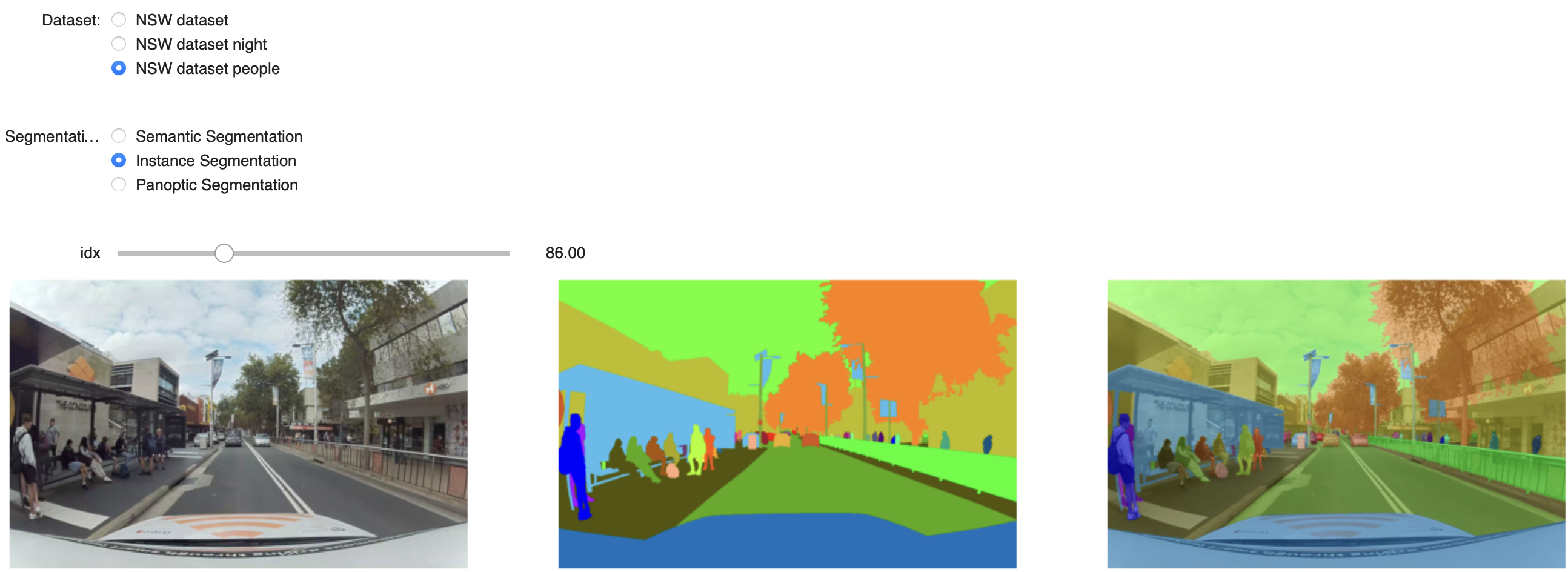}
\includegraphics[trim={32cm 0 0 8cm},clip,width=0.48\columnwidth]{figures/Visualisation.png}
\caption{\small Crowded image and panoptic mask}
\end{subfigure}

\caption{\small Visualization of segmentation masks overlaid on images from the SydneyScapes dataset: (a) Day subset, (b) Night subset, and (c) People subset.}
\label{fig:visual}
\end{figure}

Fig. \ref{fig:visual} shows three driving scenarios, from each of the datasets, with their corresponding segmentation mask. The first row shows a daytime open road with minimal surroundings, alongside a semantic segmentation mask that uses different colors to represent each of the semantic labels. The second row represents a night-time urban street, with a semantic segmentation mask. The third row corresponds to a daytime urban street with people waiting at a bus stop, accompanied by a panoptic segmentation mask, where each color represents a different instance.

\subsection{Dataset Statistics}

After labelling the anonymised images in the SydneyScapes dataset, we computed the distribution of semantic labels and instances within each subset—SydneyScapes Day, SydneyScapes Night, and SydneyScapes People—as well as across the entire dataset. 

\begin{figure}[b!]
\centering
\includegraphics[width=\columnwidth]{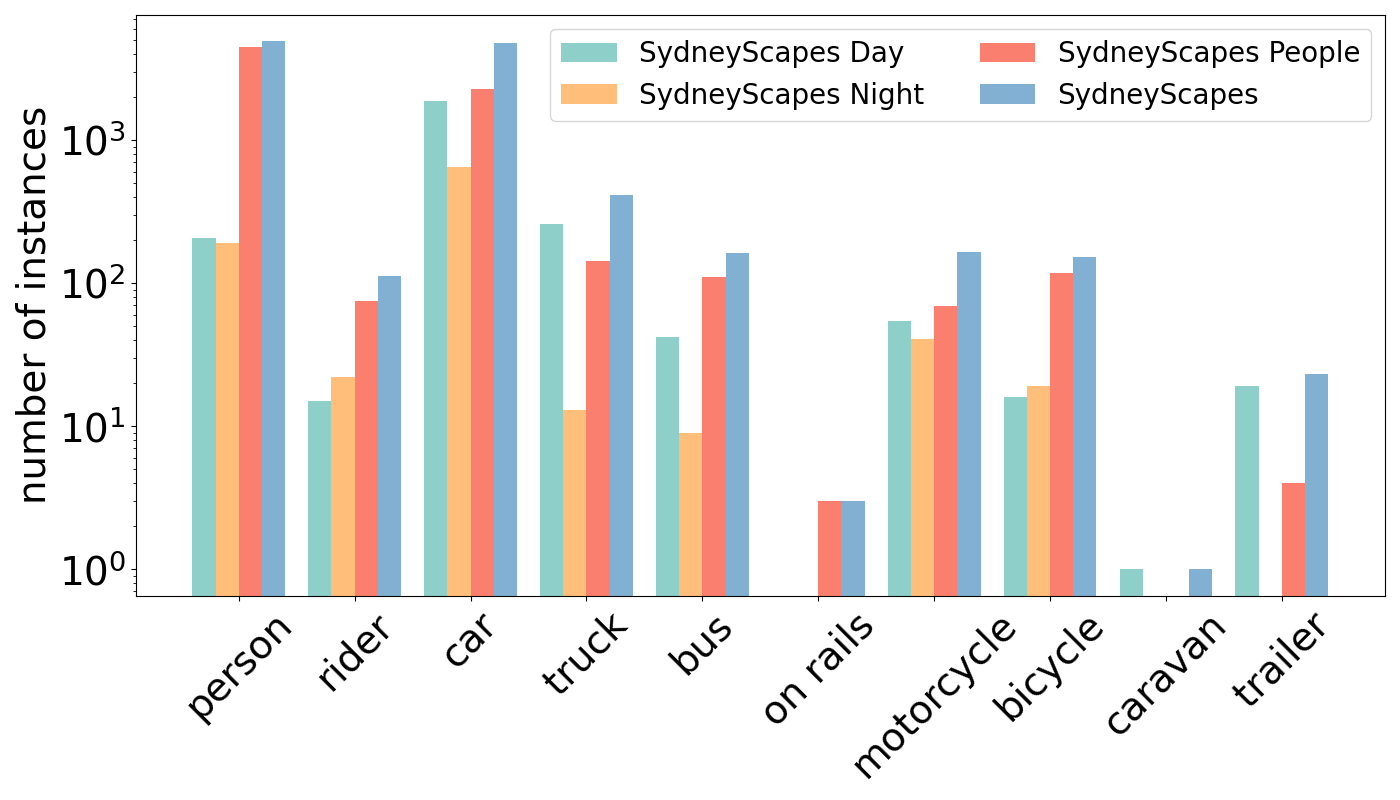}
\caption{\small Distribution of instances (y-axis) across each class (x-axis) in the SydneyScapes dataset and its subsets.}
\label{fig:instances}
\end{figure}

The bar chart in Fig. \ref{fig:semantics_distribution} shows the distribution of annotated pixels across semantic classes in the SydneyScapes dataset and its subsets. The horizontal axis lists classes such as ``road," ``building," ``sky," ``person," and ``car," among others, while the vertical axis indicates their pixel counts. This distribution shows that certain classes, like ``road," ``building," and ``sky," are prevalent across all datasets, while ``person" and ``car" are particularly prominent in the SydneyScapes People and SydneyScapes Night datasets. Other classes, such as ``bus" and ``truck," are less frequent.

For the instance segmentation task, we evaluated the number of objects for ``thing" instance classes. Fig. \ref{fig:instances} displays the distribution of instances across classes in the SydneyScapes dataset and its subsets. The vertical axis indicates instance counts and the horizontal axis lists various classes in the groups ``human" and ``vehicle". The colored bars represent different datasets: Light blue (Day) for typical driving environments, Orange (Night) collected in urban/residential environments, Red (People) focusing on pedestrian-rich areas, and Blue (Overall) representing the combined dataset.

The ``vehicle-car" class has the highest number of annotated instances, particularly in the day dataset, which was collected in typical driving environments. In the night dataset, which focuses on urban and residential areas, the ``vehicle-car" and ``human-person" classes dominate, reflecting the prevalence of cars and pedestrians in these settings. Similarly, the people dataset also shows a high concentration of ``vehicle-car" and ``human-person" instances, consistent with its urban/residential focus.

\section{Benchmarks}

In this section, we describe the benchmark evaluations on the SydneyScapes dataset for two tasks: semantic segmentation and instance segmentation. For each task, we begin by outlining the task setup and evaluation metrics, followed by the inference results of the baseline methods.
We evaluated the benchmark methods on the full validation set and subsets, categorising the results into four colour-coded sections. We first performed inference with pretrained Cityscapes weights and then fine-tuned the model with local annotations to capture environment-specific features.
 We then analyse and discuss the results of these experiments.

\subsection{Semantic Segmentation Experiments}
\subsubsection{Task and Metrics}

In image semantic segmentation, the goal is to classify each pixel into a specific category. We evaluate methods using 1928 x 1208 RGB images as input and generating class maps with labelled pixels. To align with the Cityscapes \cite{Cordts2016Cityscapes} experimental setup, we selected 19 categories from the SydneyScape dataset for assessing model performance in Australian driving scenarios.

To evaluate a semantic segmentation model, we use the mean Intersection over Union (mIoU) metric, also known as the mean Jaccard Index, which averages the Intersection over Union (IoU) \cite{everingham2015pascal} for all classes:

\begin{equation}
\text{IoU}_c = \frac{|A_c \cap B_c|}{|A_c \cup B_c|}
\end{equation}

\begin{equation}
\text{mIoU} = \frac{1}{C} \sum_{c=1}^{C} \text{IoU}_c
\end{equation}

where \( A_c \) is the set of pixels in category \( c \) from the predictions, \( B_c \) is the set of pixels in category \( c \) from the ground truth, and \( C \) is the total number of categories.

\subsubsection{Baseline Approaches}

\begin{table*}[t!]
\centering
\caption{\small \textbf{Semantic segmentation results on the SydneyScapes validation set and its subsets}. \raisebox{0pt}[0pt][0pt]{\colorbox{lightblue}{Blue}} represents results for the day subset (first and second sections). \raisebox{0pt}[0pt][0pt]{\colorbox{lightyellow}{Yellow}} represents results for the night subset (third and fourth sections). \raisebox{0pt}[0pt][0pt]{\colorbox{lightgray}{Gray}} represents results for the people subset (fifth and sixth sections). \raisebox{0pt}[0pt][0pt]{\colorbox{lightgreen}{Green}} represents results for the full validation set (seventh and eighth sections). The methods evaluated are ISANet \cite{huang2019interlaced}, BiSeNetV2 \cite{yu2021bisenet}, STDC \cite{fan2021rethinking}, SegFormer \cite{xie2021segformer}, and Mask2Former \cite{cheng2022masked}. An asterisk (*) denotes results on methods pretrained on Cityscapes \cite{Cordts2016Cityscapes}. 
A dagger $\ddagger$ depicts results on methods fine-tuned with SydneyScapes.
A hyphen (-) indicates that the field is not applicable.}
\resizebox{\textwidth}{!}{%
\begin{tabular}{l|c|ccccccccccccccccccc}
\toprule
\multicolumn{1}{c}{} & \multicolumn{1}{c}{} & \multicolumn{3}{c}{\textbf{flat}} & \multicolumn{3}{c}{\textbf{construction}} & \multicolumn{3}{c}{\textbf{object}} & \multicolumn{2}{c}{\textbf{nature}} & \multicolumn{2}{c}{\textbf{human}} & \multicolumn{6}{c}{\textbf{vehicle}} \\
\cmidrule(lr){1-1} \cmidrule(lr){2-2} \cmidrule(lr){3-5} \cmidrule(lr){6-8} \cmidrule(lr){9-11} \cmidrule(lr){12-13} \cmidrule(lr){14-15} \cmidrule(lr){16-21} 
\textbf{Method} & \rotatebox{90}{\textbf{mIoU}} & \rotatebox{90}{\textbf{road}} & \rotatebox{90}{\textbf{sidewalk}} & \rotatebox{90}{\textbf{terrain}} & \rotatebox{90}{\textbf{building}} & \rotatebox{90}{\textbf{wall}} & \rotatebox{90}{\textbf{fence}} & \rotatebox{90}{\textbf{pole}} & \rotatebox{90}{\textbf{traffic light}} & \rotatebox{90}{\textbf{traffic sign}} & \rotatebox{90}{\textbf{vegetation}} & \rotatebox{90}{\textbf{sky}} & \rotatebox{90}{\textbf{person}} & \rotatebox{90}{\textbf{rider}} & \rotatebox{90}{\textbf{car}} & \rotatebox{90}{\textbf{truck}} & \rotatebox{90}{\textbf{bus}} & \rotatebox{90}{\textbf{on rails}} & \rotatebox{90}{\textbf{motorcycle}} & \rotatebox{90}{\textbf{bicycle}} \\
\midrule
\rowcolor{lightblue} BiSeNetV2*  & 32.19 & 65.92 & 27.82 & 46.68 & 66.41 & 3.7 & 20.89 & 16.04 & 14.66 & 25.95 & 70.84 & 85.98 & 10.8 & 0.0 & 28.63 & 7.33 & 42.64 & - & 45.04 & 0.0 \\ 
\rowcolor{lightblue} ISANet*  & 28.7 & 40.15 & 19.89 & 44.81 & 37.95 & 15.66 & 30.16 & 24.25 & 19.17 & 30.98 & 54.72 & 60.4 & 9.44 & 0.0 & 47.73 & 24.18 & 5.56 & - & 51.54 & 0.0 \\
\rowcolor{lightblue} STDC*  & 29.81 & 52.58 & 22.51 & 35.51 & 45.75 & 9.65 & 24.15 & 20.3 & 14.45 & 23.88 & 63.51 & 70.55 & 8.07 & 0.23 & 33.32 & 39.77 & 16.94 & - & 55.21 & 0.19 \\
\rowcolor{lightblue} SegFormer*  & 40.94 & 77.7 & 41.09 & 61.11 & 65.12 & 3.03 & 30.85 & 30.82 & 28.44 & 40.04 & 75.35 & 90.81 & 26.7 & 0.0 & 56.14 & 35.3 & 16.49 & - & 55.84 & 2.0 \\
\rowcolor{lightblue} Mask2Former*  & 46.12 & 85.1 & 40.93 & 60.08 & 72.92 & 1.81 & 23.08 & 23.33 & 22.41 & 40.72 & 74.47 & 92.2 & 26.42 & 0.0 & 71.87 & 57.92 & 62.51 & - & 74.44 & 0.0 \\
\midrule
\rowcolor{lightblue} BiSeNetV2$\ddagger$  & 54.31 & 93.54 & 65.97 & 81.41 & 80.71 & 54.13 & 48.09 & 30.11 & 20.95 & 29.3 & 86.09 & 98.33 & 36.82 & 0.0 & 70.4 & 52.17 & 70.67 & - & 58.95 & 0.0 \\ 
\rowcolor{lightblue} ISANet$\ddagger$  & 65.46 & 94.96 & 70.63 & 85.79 & 86.48 & 69.04 & 68.61 & 42.71 & 48.43 & 66.72 & 89.35 & 98.6 & 57.54 & 5.73 & 83.21 & 73.43 & 77.62 & - & 59.48 & 0.0 \\ 
\rowcolor{lightblue} STDC$\ddagger$  & 69.13 & 95.62 & 74.32 & 86.82 & 86.0 & 71.54 & 68.52 & 46.81 & 49.27 & 61.5 & 89.41 & 98.57 & 59.63 & 30.76 & 85.57 & 73.08 & 92.57 & - & 74.28 & 0.0 \\
\rowcolor{lightblue} SegFormer$\ddagger$  & 72.18 & 96.78 & 78.66 & 86.94 & 88.34 & 72.63 & 68.23 & 54.18 & 56.87 & 62.18 & 90.9 & 99.09 & 61.71 & 26.33 & 81.2 & 78.31 & 76.25 & - & 78.79 & 41.78 \\
\rowcolor{lightblue} Mask2Former$\ddagger$  & 75.56 & 97.39 & 83.2 & 89.14 & 88.93 & 72.35 & 72.42 & 67.56 & 72.82 & 63.48 & 91.74 & 99.1 & 67.79 & 44.35 & 87.59 & 79.42 & 97.64 & - & 85.23 & 0.0 \\
\midrule
\rowcolor{lightyellow} BiSeNetV2*  & 16.68 & 48.36 & 21.68 & 0.12 & 43.26 & 0.0 & 0.0 & 9.54 & 1.75 & 21.12 & 51.77 & 14.76 & 15.79 & 22.71 & 24.22 & 0.0 & 0.0 & - & 0.0 & 25.14 \\ 
\rowcolor{lightyellow} ISANet*  & 16.2 & 24.31 & 16.99 & 6.52 & 47.31 & 0.0 & 0.0 & 19.36 & 3.09 & 9.89 & 53.16 & 9.82 & 18.17 & 0.15 & 20.29 & 0.0 & 0.0 & - & 7.42 & 55.12 \\
\rowcolor{lightyellow} STDC*  & 14.37 & 27.41 & 22.02 & 3.08 & 40.97 & 0.0 & 0.0 & 12.96 & 2.19 & 12.05 & 51.32 & 8.36 & 18.9 & 0.0 & 21.15 & 0.0 & 0.0 & - & 5.46 & 32.81 \\
\rowcolor{lightyellow} SegFormer*  & 24.5 & 65.59 & 23.6 & 0.84 & 54.7 & 0.0 & 0.0 & 20.25 & 30.99 & 19.23 & 57.98 & 13.62 & 20.73 & 27.93 & 45.98 & 0.0 & 0.0 & - & 14.36 & 45.19 \\
\rowcolor{lightyellow} Mask2Former*  & 34.23 & 81.53 & 34.14 & 0.19 & 59.53 & 0.0 & 0.05 & 27.17 & 8.07 & 36.82 & 62.47 & 24.16 & 41.94 & 45.65 & 64.67 & 1.92 & 0.0 & - & 50.18 & 77.56 \\
\midrule
\rowcolor{lightyellow} BiSeNetV2$\ddagger$  & 35.29 & 92.27 & 69.98 & 0.0 & 80.89 & 0.0 & 0.0 & 26.73 & 29.39 & 32.42 & 75.28 & 85.42 & 18.73 & 0.0 & 79.63 & 0.0 & 0.0 & - & 44.55 & 0.0 \\ 
\rowcolor{lightyellow} ISANet$\ddagger$  & 48.30 & 95.46 & 76.1 & 0.74 & 85.62 & 0.0 & 23.8 & 36.93 & 43.74 & 64.48 & 81.53 & 89.21 & 46.03 & 51.24 & 87.66 & 14.76 & 0.0 & - & 31.72 & 40.4 \\ 
\rowcolor{lightyellow} STDC$\ddagger$  & 52.62 & 95.62 & 77.69 & 1.83 & 85.6 & 0.0 & 12.23 & 35.5 & 39.78 & 54.95 & 80.64 & 88.48 & 52.26 & 37.62 & 88.41 & 8.21 & 0.0 & - & 70.92 & 67.64 \\
\rowcolor{lightyellow} SegFormer$\ddagger$  & 49.84 & 95.01 & 76.01 & 5.94 & 84.9 & 0.0 & 4.99 & 33.02 & 46.7 & 56.68 & 79.82 & 89.23 & 45.19 & 39.71 & 87.2 & 9.65 & 0.0 & - & 77.43 & 65.62 \\
\rowcolor{lightyellow} Mask2Former$\ddagger$  & 57.54 & 96.7 & 84.27 & 0.0 & 88.02 & 0.0 & 69.68 & 56.19 & 52.53 & 68.61 & 83.27 & 91.07 & 53.47 & 66.63 & 90.97 & 12.36 & 0.0 & - & 78.6 & 43.41 \\
\midrule
\rowcolor{lightgray} BiSeNetV2*  & 49.61 & 90.29 & 67.12 & 21.36 & 85.78 & 9.29 & 37.78 & 36.36 & 29.25 & 31.64 & 76.77 & 91.19 & 67.31 & 36.67 & 80.02 & 55.91 & 46.72 & 6.81 & 20.03 & 52.32 \\ 
\rowcolor{lightgray} ISANet*  & 46.30 & 70.55 & 56.52 & 37.62 & 75.72 & 9.44 & 40.13 & 41.11 & 31.03 & 37.27 & 77.82 & 86.18 & 67.63 & 39.27 & 82.32 & 30.56 & 27.73 & 3.03 & 8.65 & 57.14 \\
\rowcolor{lightgray} STDC*  & 47.25 & 82.84 & 65.84 & 27.09 & 80.14 & 7.45 & 25.12 & 35.7 & 31.86 & 30.82 & 77.34 & 88.9 & 65.4 & 34.73 & 78.07 & 32.19 & 45.65 & 24.11 & 5.23 & 59.24 \\
\rowcolor{lightgray} SegFormer*  & 54.41 & 89.0 & 63.43 & 33.78 & 88.64 & 5.22 & 42.04 & 46.75 & 51.98 & 36.52 & 81.87 & 93.94 & 74.05 & 47.36 & 85.39 & 63.77 & 41.5 & 10.07 & 15.44 & 63.04 \\
\rowcolor{lightgray} Mask2Former*  & 57.48 & 93.78 & 76.95 & 22.96 & 89.93 & 3.37 & 44.53 & 50.73 & 51.36 & 45.25 & 80.65 & 95.25 & 78.82 & 46.16 & 91.3 & 69.78 & 48.05 & 9.51 & 26.06 & 67.79 \\
\midrule
\rowcolor{lightgray} BiSeNetV2$\ddagger$  & 47.65 & 93.79 & 77.03 & 36.91 & 89.08 & 0.46 & 46.6 & 41.47 & 50.88 & 31.68 & 82.35 & 96.07 & 63.02 & 0.0 & 86.68 & 60.99 & 43.83 & 0.0 & 0.1 & 4.41 \\ 
\rowcolor{lightgray} ISANet$\ddagger$  & 65.34 & 96.28 & 85.94 & 58.66 & 93.16 & 10.12 & 61.31 & 58.76 & 76.3 & 61.74 & 88.71 & 96.67 & 77.71 & 32.65 & 93.2 & 76.52 & 78.29 & 0.0 & 28.91 & 66.48 \\ 
\rowcolor{lightgray} STDC$\ddagger$  & 64.72 & 96.58 & 85.73 & 45.35 & 93.02 & 25.35 & 61.49 & 56.11 & 73.57 & 61.92 & 88.04 & 96.59 & 73.62 & 42.75 & 90.87 & 78.9 & 78.42 & 0.0 & 21.17 & 60.23 \\
\rowcolor{lightgray} SegFormer$\ddagger$  & 67.90 & 96.91 & 87.81 & 55.11 & 94.01 & 16.84 & 66.67 & 64.43 & 79.46 & 67.24 & 89.54 & 97.92 & 78.81 & 46.49 & 94.23 & 86.53 & 76.56 & 0.03 & 22.64 & 68.87 \\
\rowcolor{lightgray} Mask2Former$\ddagger$  & 73.52 & 97.37 & 89.98 & 71.22 & 95.35 & 8.02 & 80.41 & 76.11 & 86.34 & 78.19 & 91.3 & 98.09 & 84.58 & 52.75 & 95.92 & 90.53 & 80.18 & 2.18 & 39.13 & 79.21 \\
\midrule
\rowcolor{lightgreen} BiSeNetV2*  & 40.22 & 73.84 & 45.43 & 41.96 & 72.49 & 1.86 & 23.92 & 26.84 & 25.24 & 28.78 & 66.92 & 82.5 & 60.37 & 32.13 & 48.56 & 31.87 & 40.45 & 5.99 & 20.58 & 34.45 \\ 
\rowcolor{lightgreen} ISANet*  & 38.07 & 50.34 & 39.09 & 44.6 & 58.41 & 14.4 & 32.33 & 33.9 & 27.42 & 32.43 & 59.97 & 63.45 & 60.78 & 33.22 & 59.47 & 25.4 & 6.01 & 1.65 & 25.14 & 55.28 \\
\rowcolor{lightgreen} STDC*  & 38.37 & 61.55 & 43.58 & 35.15 & 61.49 & 9.13 & 24.23 & 27.93 & 27.39 & 26.78 & 63.91 & 70.99 & 57.89 & 21.29 & 52.25 & 35.48 & 26.18 & 10.45 & 27.87 & 45.4 \\
\rowcolor{lightgreen} SegFormer*  & 48.98 & 80.76 & 49.9 & 58.93 & 75.21 & 2.73 & 32.11 & 39.21 & 47.07 & 35.53 & 73.12 & 86.42 & 70.83 & 43.59 & 69.0 & 48.66 & 26.34 & 2.82 & 32.01 & 56.4 \\
\rowcolor{lightgreen} Mask2Former*  & 54.54 & 88.16 & 57.02 & 53.15 & 79.39 & 2.02 & 24.47 & 39.01 & 43.9 & 42.99 & 73.48 & 88.39 & 75.29 & 43.02 & 80.86 & 61.03 & 52.92 & 2.95 & 59.31 & 68.87 \\
\midrule
\rowcolor{lightgreen} BiSeNetV2$\ddagger$  & 53.34 & 93.49 & 72.55 & 80.61 & 85.72 & 46.05 & 47.44 & 36.27 & 45.35 & 31.08 & 82.6 & 96.86 & 60.88 & 0.0 & 81.24 & 55.98 & 53.17 & 0.0 & 40.47 & 3.74 \\ 
\rowcolor{lightgreen} ISANet$\ddagger$  & 68.31 & 95.56 & 79.62 & 84.54 & 90.3 & 59.42 & 65.97 & 51.07 & 70.01 & 63.47 & 87.46 & 97.47 & 76.14 & 34.45 & 89.72 & 74.17 & 78.02 & 0.0 & 43.38 & 57.18 \\ 
\rowcolor{lightgreen} STDC$\ddagger$  & 70.47 & 96.01 & 81.01 & 85.62 & 90.1 & 63.22 & 65.9 & 50.52 & 67.82 & 60.97 & 87.08 & 97.38 & 72.75 & 41.83 & 89.13 & 75.18 & 83.51 & 0.0 & 69.62 & 61.22 \\
\rowcolor{lightgreen} SegFormer$\ddagger$  & 72.79 & 96.62 & 83.13 & 86.29 & 91.06 & 63.19 & 67.43 & 56.95 & 73.77 & 64.49 & 88.02 & 98.11 & 77.49 & 45.23 & 89.69 & 81.48 & 76.4 & 0.03 & 75.53 & 68.21 \\
\rowcolor{lightgreen} Mask2Former$\ddagger$  & 77.21 & 97.3 & 87.03 & 88.7 & 92.58 & 60.27 & 74.97 & 70.7 & 81.88 & 72.22 & 89.77 & 98.3 & 83.04 & 55.04 & 92.93 & 84.05 & 86.12 & 2.18 & 80.23 & 69.8 \\
\bottomrule
\end{tabular}%
}
\label{tab:semantics_benchmark}
\end{table*}

We evaluated five state-of-the-art image semantic segmentation methods on the SydneyScapes dataset: ISANet \cite{huang2019interlaced}, BiSeNetV2 \cite{yu2021bisenet}, STDC \cite{fan2021rethinking}, SegFormer \cite{xie2021segformer}, and Mask2Former \cite{cheng2022masked}.

New advancements in semantic segmentation balances accuracy and computational efficiency. The BiSeNet series \cite{yu2018bisenet, yu2021bisenet} features a dual-branch network design, where the detail branch captures fine-grained details and the semantic branch extracts high-level information. This approach looks to balance accuracy and speed, with the guided aggregation layer combining features from both branches to enhance performance. However, the additional pathways for spatial encoding might not be efficient.

To address this issue, STDC \cite{fan2021rethinking} introduces a short-term dense connection network that progressively reduces and aggregates feature map dimensions for efficient image representation. Additionally, the network uses a detail aggregation module to integrate spatial information into the low-level features, improving accuracy without increasing inference time.

Meanwhile, ISANet \cite{huang2019interlaced} improves efficiency through Interleaved Sparse Self-Attention (ISSA). ISSA decomposes the dense similarity matrix into two sparse matrices to separately estimate long-range and short-range similarities. Using ResNet \cite{he2016deep} as its backbone, ISANet applies ISSA to produce segmentation maps matching the input image's dimensions.

Recent Transformer-based approaches have advanced semantic segmentation by improving feature extraction and localisation. SegFormer \cite{xie2021segformer} presents a framework that combines Transformer and multi-layer perceptron (MLP). Its transformer encoder produces multi-scale features without positional encoding, avoiding performance issues across different resolutions. The MLP decoder aggregates multi-scale features to generate robust representations for effective segmentation.

Mask2Former \cite{cheng2022masked} presents a Masked-attention Mask Transformer for universal image segmentation. Its Transformer decoder uses masked attention to focus on predicted mask regions, enhancing feature localization and speeding up convergence. An multi-scale strategy further leverages feature maps of different resolutions to improve small object segmentation.

\subsubsection{Results and Discussion}

Table \ref{tab:semantics_benchmark} shows the semantic segmentation results for the SydneyScapes dataset. Fine-tuning the baseline methods with local annotated data significantly improves segmentation performance across the full validation set and individual subsets. The gains are largest on the day subset and smallest on the people subset. For instance, after fine-tuning, Mask2Former's mIoU increases by 22.67\% on the full validation set, 29.44\% on the day subset, 23.31\% on the night subset, and 16.04\% on the people subset. This is likely because the day subset includes images from rural areas in New South Wales, which differs from the Cityscapes dataset, while the people subset features images from crowded urban areas, resulting in a smaller domain shift.

However, fine-tuning did not improve the baseline methods' performance for all categories. For instance, Mask2Former’s IoU for the “on rails” category decreased by 0.77\% after fine-tuning. This drop may be due to the rarity of the “on rails” category, which has only about 70k annotated pixels in SydneyScapes. Fine-tuning on such a small sample for this category might introduce noise, resulting in poorer performance.

Regardless of fine-tuning, Transformer-based methods consistently outperform CNN-based methods. Mask2Former achieves the highest performance at 77.21\%, while BiSeNetV2 performs the lowest at 53.34\%. This advantage is likely due to Transformers' greater model complexity and their superior ability to encode image features through global self-attention mechanisms compared to the local receptive fields of CNNs.

\subsection{Instance Segmentation Experiments}
\subsubsection{Task and Metrics}

In instance segmentation, the goal is to predict pixel-level masks and category labels for each unique object instance. All methods use a 1928 x 1208 RGB image as input and output detailed masks and category labels for each object. In our experiments, we focused on eight categories from the ``Human" and ``Vehicle" groups, consistent with Cityscapes \cite{Cordts2016Cityscapes}.

To evaluate an instance segmentation model, we use the mean Average Precision (mAP) metric, which averages the Average Precision (AP) \cite{hariharan2014simultaneous} across all classes. AP quantifies the quality of predicted instance masks by assessing their accuracy at various IoU thresholds. Following \cite{lin2014microsoft}, we compute AP at 10 IoU thresholds ranging from 0.5 to 0.95 in 0.05 increments. We also include $\text{AP}_{50}$ for an overlap threshold of 50\% as an additional performance measure.

\subsubsection{Baseline Approaches}

We evaluated three state-of-the-art image instance segmentation methods on the SydneyScapes dataset: Mask R-CNN \cite{he2017mask}, PointRend \cite{kirillov2020pointrend}, and Mask2Former \cite{cheng2022masked}.

Mask R-CNN \cite{he2017mask} extends object detection with a parallel branch for predicting object masks. It uses RoIAlign for precise pixel-level alignment, avoiding spatial misalignment from quantiation. By decoupling classification from mask prediction, Mask R-CNN independently generates binary masks for each class, reducing class competition and improving performance.

PointRend \cite{kirillov2020pointrend} likens image segmentation to image rendering, introducing a point-based mechanism for efficient segmentation. It adaptively samples key areas (like object boundaries) non-uniformly, reducing redundant computations in smooth regions and enhancing boundary detail. The module iteratively refines uncertain regions, achieving high-resolution results with lower memory and computational costs.

Mask2Former \cite{cheng2022masked}, serving as a baseline for both semantic and instance segmentation, is compared with the previous two baselines to benchmark the performance of Transformer-based instance segmentation models on the SydneyScapes dataset.

\subsubsection{Results and Discussion}

\begin{table}[t!]
\centering
\caption{\small \textbf{Instance segmentation results on the SydneyScapes validation set and its subsets}. \raisebox{0pt}[0pt][0pt]{\colorbox{lightblue}{Blue}} represents results for the day subset (first and second sections). \raisebox{0pt}[0pt][0pt]{\colorbox{lightyellow}{Yellow}} represents results for the night subset (third and fourth sections). \raisebox{0pt}[0pt][0pt]{\colorbox{lightgray}{Gray}} represents results for the people subset (fifth and sixth sections). \raisebox{0pt}[0pt][0pt]{\colorbox{lightgreen}{Green}} represents results for the full validation set (seventh and eighth sections). The methods evaluated are Mask R-CNN \cite{he2017mask}, PointRend \cite{kirillov2020pointrend}, and Mask2Former \cite{cheng2022masked}. An asterisk (*) denotes inference on models trained on Cityscapes \cite{Cordts2016Cityscapes}. A dagger $\ddagger$ depicts results on methods fine-tuned with SydneyScapes. A hyphen (-) indicates that the field is not applicable.}
\resizebox{\columnwidth}{!}{%
\begin{tabular}{l|cc|cccccccc}
\toprule
\multicolumn{1}{l}{} & \multicolumn{2}{c}{} & \multicolumn{2}{c}{\textbf{human}} & \multicolumn{6}{c}{\textbf{vehicle}} \\
\cmidrule(lr){1-1} \cmidrule(lr){2-3} \cmidrule(lr){4-5} \cmidrule(lr){6-11}
\textbf{Method} & \rotatebox{90}{\textbf{AP}} & \rotatebox{90}{\textbf{$\text{AP}_\text{50}$}} & \rotatebox{90}{\textbf{person}} & \rotatebox{90}{\textbf{rider}} & \rotatebox{90}{\textbf{car}} & \rotatebox{90}{\textbf{truck}} & \rotatebox{90}{\textbf{bus}} & \rotatebox{90}{\textbf{on rails}} & \rotatebox{90}{\textbf{motorcycle}} & \rotatebox{90}{\textbf{bicycle}} \\
\midrule
\rowcolor{lightblue} PointRend*  & 8.3 & 16.3 & 20.3 & 0.0 & 23.6 & 4.0 & 5.6 & - & 4.3 & 0.0 \\
\rowcolor{lightblue} Mask2Former*  & 15.1 & 27.4 & 33.2 & 0.3 & 32.9 & 4.6 & 4.6 & - & 26.3 & 3.7 \\
\rowcolor{lightblue} Mask R-CNN*  & 16.4 & 27.1 & 23.4 & 0.0 & 31.1 & 10.0 & 25.0 & - & 25.0 & 0.0 \\
\midrule
\rowcolor{lightblue} PointRend$\ddagger$  & 21.4 & 37.6 & 34.9 & 0.0 & 45.2 & 25.0 & 25.0 & - & 19.5 & 0.0 \\
\rowcolor{lightblue} Mask2Former$\ddagger$  & 26.5 & 50.2 & 36.2 & 0.0 & 44.1 & 20.2 & 15.0 & - & 30.2 & 40.0 \\
\rowcolor{lightblue} Mask R-CNN$\ddagger$  & 21.0 & 33.5 & 32.4 & 0.0 & 43.5 & 21.8 & 25.0 & - & 24.6 & 0.0 \\
\midrule
\rowcolor{lightyellow} PointRend*  & 7.5 & 21.8 & 7.0 & 11.9 & 6.4 & 0.0 & - & - & 0.0 & 20.0 \\
\rowcolor{lightyellow} Mask2Former*  & 16.9 & 35.1 & 13.2 & 27.5 & 22.9 & 0.0 & - & - & 0.1 & 37.8 \\
\rowcolor{lightyellow} Mask R-CNN*  & 14.9 & 30.1 & 11.3 & 12.5 & 12.1 & 0.0 & - & - & 6.0 & 47.4 \\
\midrule
\rowcolor{lightyellow} PointRend$\ddagger$  & 26.0 & 44.3 & 22.3 & 27.5 & 47.0 & 0.0 & - & - & 26.0 & 33.3 \\
\rowcolor{lightyellow} Mask2Former$\ddagger$  & 34.3 & 64.5 & 25.4 & 25.6 & 46.3 & 25.0 & - & - & 30.3 & 53.3 \\
\rowcolor{lightyellow} Mask R-CNN$\ddagger$  & 30.3 & 60.8 & 24.3 & 25.6 & 46.6 & 7.5 & - & - & 17.1 & 60.6 \\
\midrule
\rowcolor{lightgray} PointRend*  & 16.5 & 31.0 & 26.7 & 19.2 & 39.4 & 15.2 & 18.3 & 0.0 & 1.5 & 11.7 \\
\rowcolor{lightgray} Mask2Former*  & 17.5 & 33.5 & 32.8 & 11.6 & 40.1 & 10.4 & 15.9 & 0.0 & 11.7 & 17.1 \\
\rowcolor{lightgray} Mask R-CNN*  & 19.3 & 35.7 & 32.9 & 22.0 & 39.7 & 16.4 & 20.0 & 4.2 & 3.6 & 15.3 \\
\midrule
\rowcolor{lightgray} PointRend$\ddagger$  & 24.0 & 45.7 & 35.9 & 20.0 & 46.2 & 26.0 & 35.5 & 0.0 & 11.5 & 17.0 \\
\rowcolor{lightgray} Mask2Former$\ddagger$  & 24.1 & 45.1 & 31.3 & 22.6 & 43.0 & 32.1 & 37.5 & 0.0 & 9.9 & 16.8 \\
\rowcolor{lightgray} Mask R-CNN$\ddagger$  & 25.4 & 46.9 & 37.2 & 18.6 & 45.4 & 28.0 & 43.4 & 0.0 & 13.2 & 17.3 \\
\midrule
\rowcolor{lightgreen} PointRend*  & 13.8 & 27.0 & 26.1 & 15.1 & 29.1 & 9.4 & 17.1 & 0.0 & 2.1 & 11.3 \\
\rowcolor{lightgreen} Mask2Former*  & 16.5 & 31.7 & 32.5 & 12.2 & 35.2 & 7.1 & 14.1 & 0.0 & 13.0 & 18.0 \\
\rowcolor{lightgreen} Mask R-CNN*  & 18.6 & 35.5 & 32.2 & 18.3 & 32.6 & 12.1 & 20.4 & 4.2 & 12.3 & 16.3 \\
\midrule
\rowcolor{lightgreen} PointRend$\ddagger$  & 24.1 & 45.0 & 35.6 & 19.2 & 45.9 & 24.5 & 34.6 & 0.0 & 15.4 & 17.7 \\
\rowcolor{lightgreen} Mask2Former$\ddagger$  & 24.4 & 45.8 & 31.2 & 21.2 & 43.8 & 26.9 & 35.3 & 0.0 & 17.5 & 19.4 \\
\rowcolor{lightgreen} Mask R-CNN$\ddagger$  & 25.3 & 45.8 & 36.8 & 17.9 & 44.8 & 23.4 & 41.6 & 0.0 & 17.7 & 20.3 \\
\bottomrule
\end{tabular}%
}
\label{tab:results_instace}
\end{table}

Table \ref{tab:results_instace} shows the instance segmentation results for the SydneyScapes dataset. 
Fine-tuning the baseline methods with locally labelled data significantly improves segmentation performance across the full validation set and its subsets, with the largest gains in the night subset. For instance, after fine-tuning, Mask R-CNN's AP increased by 6.7\% on the full validation set, 4.6\% on the day subset, 15.4\% on the night subset, and 6.1\% on the people subset. This is likely because the night subset's challenging lighting conditions initially lead to the poorest performance (AP of 14.9\%), leading to the largest gains from fine-tuning. 

After fine-tuning, the baseline methods show a significantly greater average AP improvement for vehicle segmentation (8.32\%) compared to human segmentation (2.1\%). The most substantial AP boost, at 21.2\%, is observed in bus segmentation. This increase is likely due to the higher variability in bus appearances in Australian traffic, as opposed to the more uniform appearance of humans globally.

\section{Conclusion}

This paper presented the SydneyScapes dataset, which offers a resource for developing and evaluating ML algorithms for Australian urban environments. This dataset features high-quality annotations for semantic, instance, and panoptic segmentation, which addresses the unique challenges presented by Australian landscapes, including diverse weather, lighting conditions, and urban settings. The benchmark results demonstrate the importance of fine-tuning models to local contexts, as the performance of segmentation methods significantly improves when adapted to the specific characteristics of the environment.

Furthermore, the availability of a user-friendly online visualisation tool enhances the dataset's accessibility and usability, supporting researchers and industry in advancing the state of autonomous vehicle perception systems in Australia. The SydneyScapes dataset, therefore, not only contributes to the global research community but also lays the groundwork for further innovations in understanding and navigating Australian scenes.

\bibliographystyle{apalike}
\bibliography{acra}

\end{document}